%% file: main.tex
\documentclass{article}

 \usepackage[preprint]{main}

\usepackage[utf8]{inputenc} 
\usepackage[T1]{fontenc}    
\usepackage{hyperref}       
\usepackage{url}            
\usepackage{booktabs}       
\usepackage{amsfonts}       
\usepackage{nicefrac}       
\usepackage{microtype}      
\usepackage{xcolor}         
\usepackage{graphicx}
\usepackage{amsmath}
\usepackage{multirow}    
\usepackage{wrapfig}
\usepackage{calc}
\usepackage{booktabs}
\usepackage[table]{xcolor} 
\usepackage{colortbl}
\usepackage{caption}
\usepackage{capt-of}
\usepackage{makecell}
\usepackage[table]{xcolor} 
\usepackage{booktabs}      
\usepackage{multirow}      
\usepackage{graphicx}      
\usepackage{hhline}        
\usepackage{xcolor}
\usepackage{enumitem}
\usepackage{bbding}

\title{HarmoWAM: Harmonizing Generalizable and Precise Manipulation via Adaptive World Action Models}

\author{
Qiuxuan Feng\thanks{Equal contribution, $^{\dagger}$Project lead, \textsuperscript{\Envelope}Corresponding author.} \hspace{0.1mm} \textsuperscript{\rm 1} \hspace{0.1mm}
Jiale Yu$^{*}$\textsuperscript{\rm 1} \hspace{0.1mm}
Jiaming Liu$^{*, \dagger}$\textsuperscript{\rm 1} \hspace{0.1mm}
Yueru Jia$^{*}$\textsuperscript{\rm 1} \hspace{0.1mm} 
Zhuangzhe Wu\textsuperscript{\rm 1} \hspace{0.1mm} \\
\textbf{Hao Chen\textsuperscript{\rm 3}} \hspace{0.1mm}
\textbf{Zezhong Qian\textsuperscript{\rm 1}} \hspace{0.1mm}
\textbf{Shuo Gu\textsuperscript{\rm 2}} \hspace{0.1mm}
\textbf{Peng Jia\textsuperscript{\rm 2}} \hspace{0.1mm}
\textbf{Siwei Ma\textsuperscript{\rm 1}} 
\hspace{0.1mm}
\textbf{Shanghang Zhang\textsuperscript{\rm 1}~\textsuperscript{\Envelope}} \hspace{0.1mm} \\
\textsuperscript{\rm 1}State Key Laboratory of Multimedia Information Processing, School of Computer Science, \\ Peking University
\textsuperscript{\rm 2}Simplexity Robotics 
\textsuperscript{\rm 3}The Chinese University of Hong Kong \vspace{0.2cm}\\
Project page: \url{https://elbb-yu.github.io/HarmoWAM/}
}

\begin{document}

\maketitle

\input{sec/1_abs}

\input{sec/2_intro}

\input{sec/3_related}

\input{sec/4_method}

\input{sec/5_experiments}

\section{Conclusion}
In this work, we identify a fundamental trade-off in existing World Action Models (WAMs) between generalizable motion transit and precise manipulation, arising from their distinct modeling paradigms. Therefore, we propose HarmoWAM, an end-to-end WAM that unifies predictive and reactive control under a shared world model. By leveraging spatio-temporal priors and coordinating two complementary action experts via a Process-Adaptive Gating Mechanism, HarmoWAM enables stage-aware control that jointly achieves broad exploration and fine-grained interaction within a unified closed-loop policy.
Extensive real-world experiments across diverse tasks and unseen environments demonstrate that HarmoWAM consistently outperforms prior approaches, achieving strong zero-shot generalization while maintaining high manipulation accuracy.

\newpage
\bibliographystyle{plain} 
\bibliography{main}


\newpage
\appendix

\noindent{\large\bfseries Appendix}
\input{sec/7_appendix}


\end{document}

%% file: sec/1_abs.tex
\begin{abstract}

World Action Models (WAMs) have emerged as a promising paradigm for robot control by modeling physical dynamics.
Current WAMs generally follow two paradigms: the "Imagine-then-Execute" approach, which uses video prediction to infer actions via inverse dynamics, and the "Joint Modeling" approach, which jointly models actions and video representations.
Based on systematic experiments, we observe a fundamental trade-off between these paradigms: the former explicitly leverages world models for generalizable transit but lacks interaction precision, whereas the latter enables fine-grained, temporally coherent action generation but is constrained by the exploration space of the training distribution.
Motivated by these findings, we propose \textbf{HarmoWAM}, an end-to-end WAM that fully leverages a world model to unify predictive and reactive control, enabling both generalizable transit and precise manipulation.
Specifically, the world model provides spatio-temporal physical priors that condition two complementary action experts: a \emph{predictive expert} that leverages latent dynamics for iterative action generation, and a \emph{reactive expert} that directly infers actions from predicted visual evolution.
To enable adaptive coordination, a Process-Adaptive Gating Mechanism is proposed to automatically determine the timing and location of switching between them.
This allows the world model to drive the reactive expert to expand the exploration space and the predictive expert to perform precise interactions across different stages of a task.
For evaluation, we construct three training-unseen test environments across six real-world robotic tasks, covering variations in background, position, and object semantics.
Notably, HarmoWAM achieves strong zero-shot generalization across these scenarios, significantly outperforming prior state-of-the-art VLA models and WAMs by margins of 33\% and 29\%, respectively.

\end{abstract}

%% file: sec/2_intro.tex
\section{Introduction}

\begin{figure*}[t] 
    \centering
    \includegraphics[width=\textwidth]{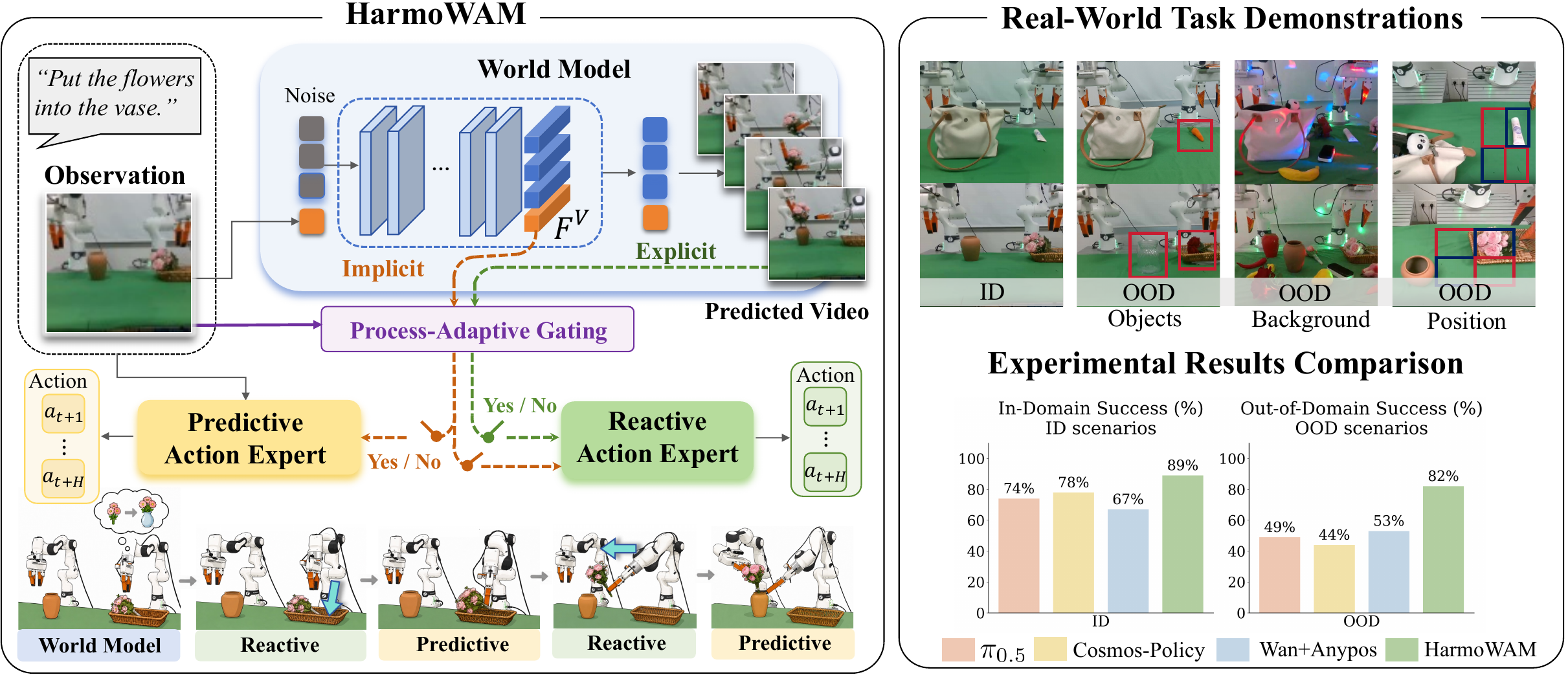} 
    \caption{\textbf{Overview.} 
    We propose HarmoWAM, an end-to-end WAM that jointly achieves generalizable transit and precise manipulation through a world model that provides physical dynamics priors and adaptively coordinates a predictive action expert and a reactive action expert. HarmoWAM achieves SOTA performance in ID settings and exhibits a substantial advantage in OOD scenarios.
    }
    \label{fig:teasor}
    \vspace{-0.3cm}
\end{figure*}

Robust robotic manipulation requires policies that not only understand task instructions but also capture the underlying physical dynamics of the environment.
\textbf{World Action Models (WAMs)} have recently emerged as a promising paradigm toward this goal, leveraging future visual prediction to provide spatio-temporal physical priors for action generation~\cite{kim2026cosmos,jang2025dreamgenunlockinggeneralizationrobot,yuan2026fastwamworldactionmodels}.
Existing WAMs can be broadly categorized into two paradigms~\cite{yuan2026fastwamworldactionmodels}: (1) \textbf{Imagine-then-Execute}, which first generates imagined future trajectories and then infers actions via inverse dynamics~\cite{feng2025vidar,jang2025dreamgenunlockinggeneralizationrobot,chi2025wow}; and (2) \textbf{Joint Modeling}, which directly models the joint distribution of actions and video representations, leveraging latent features from world models to guide action prediction~\cite{kim2026cosmos,bi2025motus,zhu2025unifiedworldmodelscoupling}. However, we observe that these two paradigms exhibit markedly different behaviors across real-world tasks.

Driven by the inherent differences between these paradigms, we conduct a systematic empirical study on two representative real-world manipulation tasks, including dual-arm collaboration and precision stacking.
For each task, we decompose the process into a transit phase between objects and an interaction phase, and evaluate performance under both In-Domain (ID) and three Out-of-Domain (OOD) settings, covering variations in background, object position, and object instances.
We benchmark two representative approaches: Imagine-then-Execute (Wan2.2-TI2V-5B + AnyPos, following~\cite{chi2025wow,tan2025anypos}) and Joint Modeling (Wan2.2-TI2V-5B + Action DiT, following~\cite{jia2026video2actdualsystemvideodiffusion,hu2025videopredictionpolicygeneralist}).
Further details are provided in Section~\ref{sec:moti}.
We observe a fundamental trade-off between these paradigms: (1) \textbf{Imagine-then-Execute} leverages the generalization capability of world models, achieving near-perfect success in object approach even under OOD conditions, but lacks the precision required for fine-grained manipulation (with average manipulation success dropping to as low as 60\%); (2) \textbf{Joint Modeling} achieves stable control precision above 80\% when initialized near target objects, but fails to reliably approach them in OOD settings (with success rates below 60\%), highlighting an exploration space constrained by the supervised fine-tuning (SFT) data distribution.

Motivated by these findings, we introduce \textbf{HarmoWAM}, an end-to-end WAM that unifies generalizable transit and precise manipulation (as illustrated in Figure~\ref{fig:teasor}). 
In this design, a world model generates future visual observations and latent representations as spatio-temporal conditions to adaptively coordinate predictive and reactive control across two complementary action experts.
Both action experts are built upon Transformer-based backbones. 
\textbf{A predictive action expert} leverages latent video features to guide fine-grained, temporally coherent action generation. Meanwhile, \textbf{a reactive expert} directly infers actions from predicted future observations and corresponding latent features.
Together, these components form a unified framework that effectively reconciles generalization and precision for robust manipulation in dynamic environments.
To enable adaptive coordination, we propose a Process-Adaptive Gating Mechanism that dynamically determines when and where to switch between two action experts based on interaction events.
Specifically, following a keyframe extraction strategy~\cite{shridhar2023perceiver}, we automatically construct a process-aware dataset to train a gating network that predicts the current task stage (i.e., transit or interaction) from visual observations.
This enables HarmoWAM to adaptively switch between the two experts across different task stages, with the world model guiding the reactive expert to expand the exploration space and the predictive expert to perform precise interactions, thereby establishing a generalizable closed-loop control framework.

To comprehensively evaluate HarmoWAM's efficacy, we conduct experiments across six real-world tasks (four single-arm, two dual-arm) under ID and three distinct OOD scenarios. As shown in Figure~\ref{fig:teasor}, HarmoWAM establishes a new state-of-the-art (SOTA) by outperforming the leading VLA and WAMs methods by 15\% and 11\% in ID settings, respectively. This performance gap further widens in OOD scenarios, where HarmoWAM demonstrates strong zero-shot generalization, outperforming state-of-the-art VLA and WAM methods by 33\% and 29\% on average across all variation types.
For inference, HarmoWAM achieves an action generation speed of 48 Hz with an action chunk size of 12.
We further extend the task horizon in real-world deployments, demonstrating that HarmoWAM maintains generalizable transit alongside high-precision manipulation even in long-horizon tasks. 
Our main contributions are as follows:

\begin{enumerate}[leftmargin=15pt, labelsep=5pt, label=\arabic*)]
\item We propose HarmoWAM, a novel end-to-end WAM that leverages spatio-temporal conditions provided by a world model to unify predictive action generation and reactive execution, effectively reconciling generalization and precision in robotic manipulation.

\item We introduce a Process-Adaptive Gating Mechanism that enables stage-aware coordination in HarmoWAM, dynamically routing control under world model guidance, the reactive expert expands the exploration space, while the predictive expert ensures precise manipulation.
\item We conduct extensive real-world evaluations across diverse manipulation tasks and OOD scenarios, showing that HarmoWAM not only achieves strong zero-shot generalization but also maintains robust performance in long-horizon settings.

\end{enumerate}

%% file: sec/3_related.tex
\section{Related Work}
\textbf{Vision-language-action (VLA)} models~\cite{brohan2022rt,2024_9_05_OpenVLA,chen2024spatialvlm,liu2026last,li2024cogact,lin2025onetwovla,pertsch2025fast} adapt pretrained vision-language models (VLMs)~\cite{2024_2_12_Prismatic_VLMs,2022_11_15_Flamingo} to embodied control, allowing robots to execute tasks based on natural language instructions. Early efforts in this line of work mainly emphasized scaling robot demonstration data and transferring pretrained VLM knowledge to action learning~\cite{zitkovich2023rt, team2024octo,li2024manipllm,gu2025manualvla}.  Building on this foundation, more recent VLA research has shifted toward strengthening the action generation module. One prominent direction replaces discrete or deterministic heads with continuous generative policies for precise control~\cite{wen2024diffusion, wen2025tinyvla, 2025_5_1_RDT_1B, 2025_3_13_HybridVLA, kim2025fine, 2025_1_16_FAST, 2025_6_02_Fast_in_Slow}. Specifically, flow-based formulations have emerged as a more efficient action generation, reducing sampling overhead while maintaining strong action fidelity~\cite{2024_10_31_pi0, 2025_4_22_pi0_5, liu2026last, intelligence2026pi}. However, despite these advances, current VLA models remain heavily constrained by their training data distribution, with generalization requiring costly data.

\textbf{World Action Models (WAMs)} have increasingly explored video generation and predictive visual modeling for robot control~\cite{zhou2025act2goal, bharadhwaj2024gen2act,ye2026gigaworld,du2023learning,wu2023unleashing,zhou2024robodreamer,won2025dual,cheang2024gr,cen2025rynnvla,cen2025worldvla,zheng2025flare}. These methods can be broadly divided into two paradigms. The first adopts an \emph{Imagine-then-Execute} strategy, where the model first predicts future visual observations and then derives actions from the imagined future frames~\cite{feng2025vidar}. In such approaches, video prediction explicitly captures environment dynamics and provides foresight for downstream decision-making, such as DreamGen~\cite{jang2025dreamgenunlockinggeneralizationrobot} and WoW~\cite{chi2025wow}. 
The second paradigm is \emph{Joint Modeling}, which jointly models video and action generation~\cite{kim2026cosmos,pai2025mimic,li2026causal}. 
Specifically, these methods incorporate predictive visual modeling into policy learning as latent supervision or action conditioning. For example, VPP~\cite{hu2025videopredictionpolicygeneralist}, Video2Act~\cite{jia2026video2actdualsystemvideodiffusion}, Genie Envisioner~\cite{liao2025genieenvisionerunifiedworld}, and Video Policy~\cite{liang2025videogeneratorsrobotpolicies} use predictive representations extracted from video diffusion models to condition action generation. 
UVA~\cite{li2025unifiedvideoactionmodel}, UWM~\cite{zhu2025unifiedworldmodelscoupling}, Cosmos policy~\cite{kim2026cosmos}, and Motus~\cite{bi2025motus} jointly model video and action generation within unified generative frameworks, while Fast-WAM~\cite{yuan2026fastwamworldactionmodels} shows that the benefits of video modeling can persist even without explicit future generation at inference time. 
Recent world action models continue to advance this trend by treating joint video-action modeling itself as a scalable paradigm for robot policy learning and generalization~\cite{ye2026worldactionmodelszeroshot}. 
In this paper, we propose HarmoWAM, an end-to-end WAM that fully exploits both explicit and implicit spatio-temporal priors from a world model to unify generalizable transit and precise manipulation.

%% file: sec/4_method.tex
\section{Methods}

\subsection{Preliminaries}
\label{sec:preliminaries}
\textbf{Action Prediction Problem Formulation.}
We formulate robotic manipulation as a probabilistic sequential decision-making problem~\cite{2024_10_31_pi0}. 
At each timestep $t$, conditioned on a natural language instruction $\mathbf{l}_{t}$ and a visual observation $\mathbf{I}_{t} \in \mathbb{R}^{H \times W \times 3}$, the policy $\pi_\theta$ predicts a future action sequence of length $H$, defined as $\mathbf{a}_{t+1:t+H} \sim \pi_\theta(\cdot \mid \mathbf{I}_t, \mathbf{l}_t)$. 
For single-arm manipulation, we employ a 7-DoF action space $\mathbf{a} \in \mathbb{R}^7$, comprising a 3-DoF relative positional displacement $(x, y, z)$, a 3-DoF Euler angle rotation $(\mathrm{roll}, \mathrm{pitch}, \mathrm{yaw})$, and a 1-DoF binary gripper state $g \in {0,1}$.
For dual-arm settings, we extend this to a 14-DoF space by concatenating the dual arm control vectors. Additional Inverse Dynamics Model (IDM) based action inference preliminaries are provided in Appendix~\ref{app:preli}.

\subsection{Motivation}
\label{sec:moti}

While existing WAMs, whether based on the Imagine-then-Execute or Joint Modeling paradigm, have demonstrated promising capabilities, our real-world experiments reveal that none can robustly handle diverse tasks. Therefore, we conduct systematic real-world evaluations on representative baselines to characterize their manipulation boundaries and uncover their inherent trade-offs.

\input{tables/motivation_elbb}

\textbf{Experimental Setup.} 
For the \emph{Imagine-then-Execute}, following~\cite{chi2025wow, mi2026tc}, we instantiate the baseline using Wan2.2-TI2V-5B~\cite{wan2025wan} for video generation and AnyPos~\cite{tan2025anypos} for action inference.
For the \emph{Joint Modeling}, following VPP~\cite{hu2025videopredictionpolicygeneralist}, we reimplement the baseline for fair comparison by integrating latent features from Wan2.2-TI2V-5B into an Action Diffusion Transformer (DiT), where video representations are introduced as conditioning signals via cross-attention. We conduct the motivation study on two representative real-world manipulation tasks: \textit{Put Flowers in Vase}, which requires long-horizon dual-arm coordination, and \textit{Stack Coke Cans}, which demands precise, fine-grained stacking manipulation. Both tasks are evaluated under in-domain (ID) settings and three out-of-domain (OOD) scenarios, covering variations in background, object position, and object semantics.

\textbf{Findings and Analysis.}
In Table~\ref{tab:motivation}, the \emph{Imagine-then-Execute} baseline achieves near-perfect success in the transit phase across both ID and OOD settings, demonstrating strong generalization in object-to-object motion. However, its performance in the interaction phase drops significantly, with average success rates below 75\% in ID and 55\% in OOD scenarios, indicating a lack of precision in fine-grained object manipulation. In contrast, the \emph{Joint Modeling} baseline achieves strong performance in ID settings, with an overall success rate of over 90\% across both transit and interaction phases. However, its transit success degrades substantially in OOD scenarios, dropping to 32\%, which reveals a limited exploration space constrained by the supervised fine-tuning (SFT) data distribution. Notably, when initialized near the target object, it still achieves an average interaction success of 95\% in OOD settings, suggesting that its limitation primarily lies in exploration rather than manipulation.
These results reveal a fundamental trade-off between the two paradigms: \emph{Imagine-then-Execute} excels at generalizable transit but lacks manipulation precision, whereas \emph{Joint Modeling} achieves high precision but fails to reliably explore and reach target objects in unseen environments.
This observation highlights the necessity of a unified WAM framework that can jointly achieve generalizable transit and precise manipulation. More details are provided in Appendix~\ref{app:motivation}.

\subsection{HarmoWAM Framework}
\label{sec:Harmo}
Motivated by the above findings, we propose HarmoWAM, a novel WAM that jointly achieves generalizable transit and precise manipulation. Going beyond architectural refinements, our design represents a first step toward a unified paradigm that organically integrates these complementary capabilities, fully unlocking the knowledge encoded in world models.

\begin{figure*}[t] 
    \centering
    \includegraphics[width=0.99\textwidth]{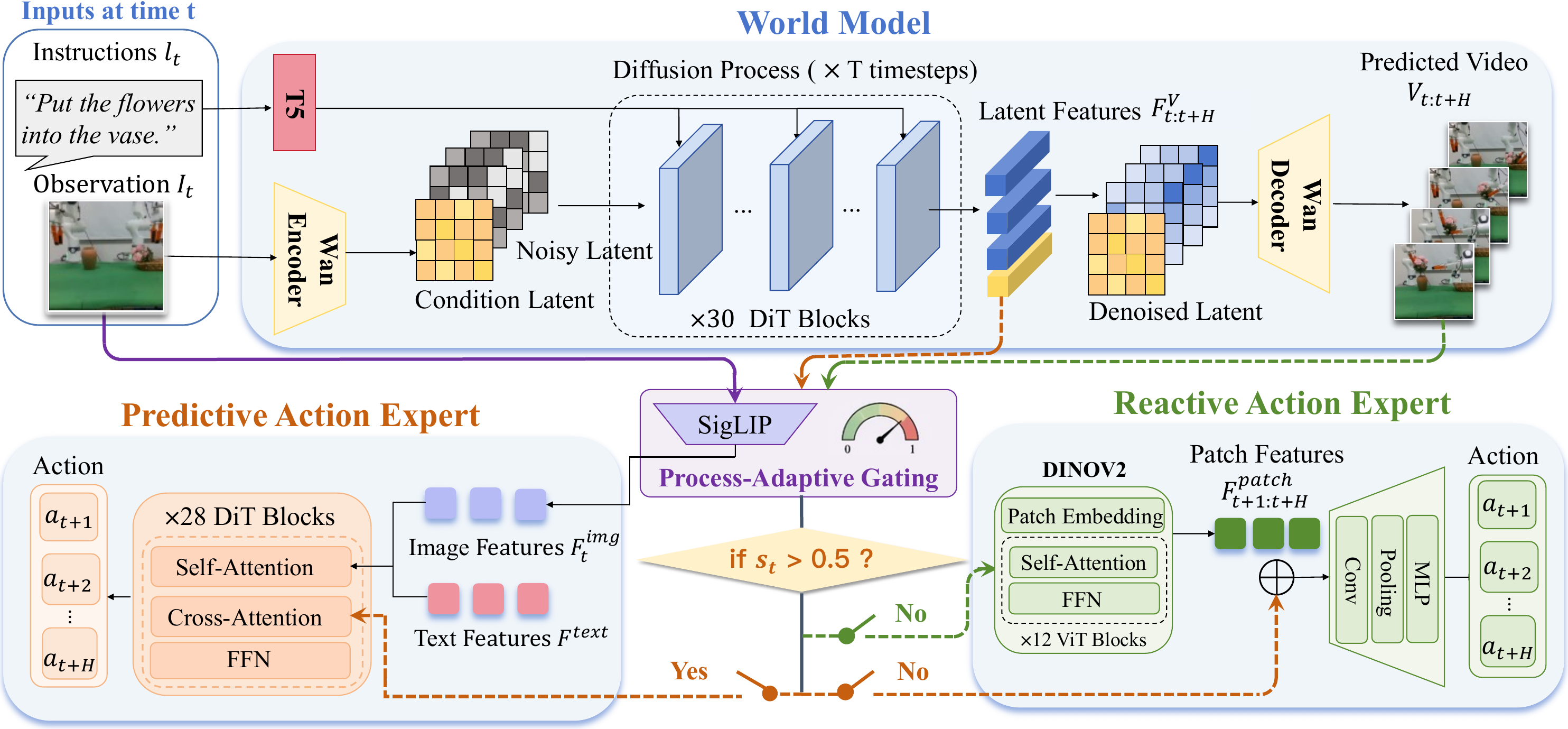} 
    \vspace{-0.1cm}
    \caption{\textbf{Framework.} 
    HarmoWAM adopts an adaptive framework that tightly integrates a generative world model with two complementary action experts. The world model provides both explicit future predictions and implicit latent representations. Conditioned on current latent features, the predictive expert generates structured actions for precise manipulation, while the reactive expert leverages future predicted frames and their latent features to perform reactive inference for generalizable transit. A Process-Adaptive Gating mechanism predicts the task stage from observations and dynamically routes control between the two action experts, resulting in a unified closed-loop control paradigm.
    }
    \vspace{-0.6cm}
    \label{fig:method}
\end{figure*}

\textbf{Overall Architecture.}
As shown in Figure~\ref{fig:method}, HarmoWAM adopts an adaptive architecture by leveraging a world model to seamlessly bridge spatio-temporal reasoning and motion generation with two complementary action experts.
The world model is instantiated using the Wan2.2-TI2V-5B~\cite{wan2025wan}, and is further pretrained on large-scale robotic data (approximately 1.9M trajectories)~\cite{bu2025agibot, wu2024robomind, lightewm2026} to endow it with robotic physical modeling and generalization capabilities. The details are shown in Appendix~\ref{app:trainingdetails}. It operates at a resolution of $256 \times 320$ and predicts a 13-frame future video for each input observation. 
To ground the world model's physical dynamics into actionable robot control, we design two complementary experts. Both action experts are built upon Transformer-based backbones. 
The predictive expert is instantiated as a 1B-parameter DiT, consisting of 28 Transformer blocks and incorporating a SigLIP~\cite{zhai2023sigmoid} image encoder and a text encoder~\cite{raffel2020exploring} to provide real-time, closed-loop visual observations and task instructions, enabling structured action generations. 
Meanwhile, the reactive expert is designed for real-time and interpretable action generation by directly leveraging the generalizable visual predictions from the world model. It integrates a DINOv2-base~\cite{oquab2023dinov2} model with an Orientation Decoder ($\mathcal{D}_{ori}$), where the decoder is implemented using multi-scale convolutional layers~\cite{tan2025anypos} to map high-level visual features into the action space.

\textbf{Coordination in HarmoWAM.}
Unlike prior approaches that treat world models and action policies as loosely coupled modules, HarmoWAM introduces a fully differentiable framework that tightly integrates a world model with tailored action experts.
We explicitly design complementary experts to effectively exploit both the implicit latent representations and explicit video predictions generated by the world model.
Given the current visual observation and instruction, the world model predicts future visual evolution $\mathbf{V}_{t:t+H}$ via a video diffusion process (5 denoising steps), and extracts latent representations $\mathcal{F}_{t:t+H}^{\mathbf{V}}$ as spatio-temporal conditions. 

\textit{World Model to Predictive Expert.}
To fully exploit latent representations from the world model, which encode rich temporal dynamics critical for action generation, the predictive expert leverages the current-step latent feature $\mathcal{F}^{\mathbf{V}}_{t} \in \mathbb{R}^{B \times 80 \times 3072}$ as an implicit condition.
This feature is incorporated into the Action DiT through cross-attention layers, together with image features $\mathcal{F}^{img}_{t}$ and textual features $\mathcal{F}^{text}$. 
At each action diffusion denoising step $k$, the model predicts the added Gaussian noise $\epsilon_\theta = \mathcal{D}_{\theta_{\mathrm{pred}}}
({\mathbf{a}}_{t+1:t+H}, \tau_k \mid 
\mathcal{F}^{img}_{t}, \mathcal{F}^{text}, \mathcal{F}^{\mathbf{V}}_{t})$, where $\tau_k$ denotes the timestep embedding. This temporally coherent conditioning enables the expert to internalize physical dynamics during action prediction, leading to more structured and consistent decision-making.

\textit{World Model to Reactive Expert.}
Unlike prior IDM approaches that rely solely on observed video frames, which primarily provide low-level visual cues, we additionally incorporate the corresponding future latent dynamics. 
Concretely, for each future timestep $s \in \{t+1,\dots,t+H\}$, the reactive expert takes both the predicted frame $\mathbf{V}_{s}$ and its latent representation $\mathcal{F}^{\mathbf{V}}_{s} \in \mathbb{R}^{B \times 80 \times 3072}$ as inputs. 
A DINOv2 encoder extracts patch-level geometric features $\mathcal{F}^{\text{patch}}_{s} \in \mathbb{R}^{B \times 1369 \times 768}$ from $\mathbf{V}_{s}$. 
These features are concatenated with the latent features along the token dimension, after applying average pooling to reduce the channel dimension of $\mathcal{F}^{\mathbf{V}}_{s}$ from 3072 to 768, forming a fused representation:
$\mathcal{F}^{\text{fuse}}_{s} = 
    [ \mathcal{F}^{\text{patch}}_{s} ; {\mathcal{F}}^{\mathbf{V}}_{s} ] .$
The Orientation Decoder $\mathcal{D}_{\text{ori}}$ extracts a global representation to predict the action:
$
\hat{\mathbf{a}}_{s} = \mathcal{D}_{\text{ori}}(\mathcal{F}^{\text{fuse}}_{s}).
$
This design allows inverse dynamics to be performed not only based on low-level visual observations but also informed by high-level spatio-temporal knowledge.

\subsection{\textbf{Process-Adaptive Gating Mechanism}}
\label{sec:gating}
To ensure efficient synergy within the HarmoWAM framework, we propose a Process-Adaptive Gating Mechanism. 
This mechanism is motivated by the transit-interaction complementarity observed in Section~\ref{sec:moti} and is designed to evaluate observations in real time and dynamically manage the switching logic between the two action experts. 
Specifically, it is implemented as a lightweight MLP-based classification network. 
We reuse the visual tokens $\mathcal{F}^{img}_{t}$ from the current observation as inputs to the gating network.
To estimate whether the current task state favors precise interaction or generalizable transit, the network applies a non-linear mapping to output a confidence score $s_t \in [0, 1]$.
During supervised training, we follow a keyframe extraction pipeline~\cite{shridhar2023perceiver} to automatically construct a process-aware dataset from robot proprioceptive signals.
Interaction phases are identified by changes in gripper state or task-specific end-effector height thresholds, which indicate moments requiring precise manipulation, and are labeled as $y=1$ for the predictive action expert.
In contrast, frames without such interaction cues correspond to transit phases (e.g., approaching the target) and are labeled as $y=0$ for the reactive action expert.
More details are provided in Appendix~\ref{app:gating_details}. The gating network is optimized using a Binary Cross-Entropy (BCE) loss function $\mathcal{L}_{gate}$ as follows:\begin{equation}\mathcal{L}_{gate} = -\frac{1}{N} \sum_{i=1}^{N} \left[ y_i \log(s_i) + (1 - y_i) \log(1 - s_i) \right]\end{equation} where $N$ is the batch size, $y_i \in {0,1}$ is the label for frame $i$, and $s_i \in [0,1]$ is the predicted probability of it being in an interaction phase.
During inference, an activation threshold is set to $0.5$. When the confidence score $s_t > 0.5$, the gating mechanism routes the world model’s representations to the predictive expert for action inference. When $s_t \leq 0.5$, the system leverages reactive expert to expand the exploration space. This allows HarmoWAM to dynamically select the appropriate action expert, thereby maintaining both generalization and precision throughout the manipulation process. 
To justify our design, we visualize the attention maps of the two complementary experts in the upper part of Figure~\ref{fig:real_execution}. We observe that the predictive expert focuses more on manipulated objects, while the reactive expert attends to the robot gripper and task-relevant surrounding environment.

\subsection{Downstream Training Recipe}
To effectively align the learning of physical dynamics with downstream action generation, HarmoWAM is optimized through a two-stage training paradigm.

\textbf{Stage 1: World Model Finetuning.}
We first adapt a pretrained world model to capture task-specific visual dynamics from real-world demonstrations.
The model is fully fine-tuned with a conditional Flow Matching objective.
Let $\mathbf{x}_1$ denote the clean video latent from a demonstration and $\mathbf{x}_0 \sim \mathcal{N}(0,I)$ denote a Gaussian noise latent.
Given the condition $\mathbf{c}$, including the current observation and task instruction, we sample a flow interpolation variable $\xi \in [0,1]$ and construct:
$\mathbf{x}_\xi = (1-\xi)\mathbf{x}_0 + \xi\mathbf{x}_1$.
The target velocity is defined as:
$\mathbf{v}_\xi = \frac{d\mathbf{x}_\xi}{d\xi} = \mathbf{x}_1-\mathbf{x}_0$.
The training loss is:
\begin{equation}
\mathcal{L}_{\mathrm{stage1}} =
\mathbb{E}_{\mathbf{x}_0,\mathbf{x}_1,\xi,\mathbf{c}}
\left[
w(\xi)
\left\|
f_\theta(\mathbf{x}_\xi,\xi,\mathbf{c}) - \mathbf{v}_\xi
\right\|_2^2
\right],
\end{equation}
where $w(\xi)$ is the flow-step-dependent weighting function.

\textbf{Stage 2: Action Experts Finetuning.} 
We optimize the two experts and the gating network, conditioned on the explicit and implicit video features provided by the frozen world model.
\textbf{The predictive expert} is trained using a standard diffusion denoising loss ($\mathcal{L}_{\mathrm{pred}}$) to ensure accurate action generation, where $\epsilon$ denotes the Gaussian noise added to the action sequence.
\textbf{The reactive expert} is supervised with a Smooth L1 loss ($\mathcal{L}_{\mathrm{react}}$) to align the predicted action $\hat{\mathbf{a}}_t$ with the expert demonstration action $\mathbf{a}_t$.
\begin{equation}
\mathcal{L}_{\mathrm{pred}} = \mathbb{E}_{\mathbf{a}_{t+1:t+H}, \epsilon \sim \mathcal{N}(0,1)} \left[ \| \epsilon_\theta - \epsilon \|_2^2 \right], \quad
\mathcal{L}_{\mathrm{react}} = \mathbb{E} \left[ d(\hat{\mathbf{a}}_{t+1:t+H}, \mathbf{a}_{t+1:t+H}) \right].
\end{equation}

Integrating these sub-objectives alongside the gating loss $\mathcal{L}_{\mathrm{gate}}$ defined previously, the overall training objective is formulated as a weighted sum:
\begin{equation}
\label{eq:total_loss}
\mathcal{L}_{\mathrm{stage2}} =  \mathcal{L}_{\mathrm{pred}} + \lambda_{react} \mathcal{L}_{\mathrm{react}} + \lambda_{\mathrm{gate}} \mathcal{L}_{\mathrm{gate}}
\end{equation}
where $\lambda_{react} = 0.1$ and $\lambda_{\mathrm{gate}} = 0.05$ are used to balance the scales of different losses.

%% file: tables/motivation_elbb.tex
\begin{table}[t]
  \centering
  \vspace{-0.1cm}
  \caption{\textbf{Motivation.} Comparison of \emph{Imagine-then-Execute} and \emph{Joint Modeling} under ID and OOD settings. Interaction performance ($^*$) is measured with the robot initialized near the target object.}
  \label{tab:motivation}
  
  \begin{minipage}[c]{0.43\linewidth}
    \centering
    \includegraphics[width=\linewidth]{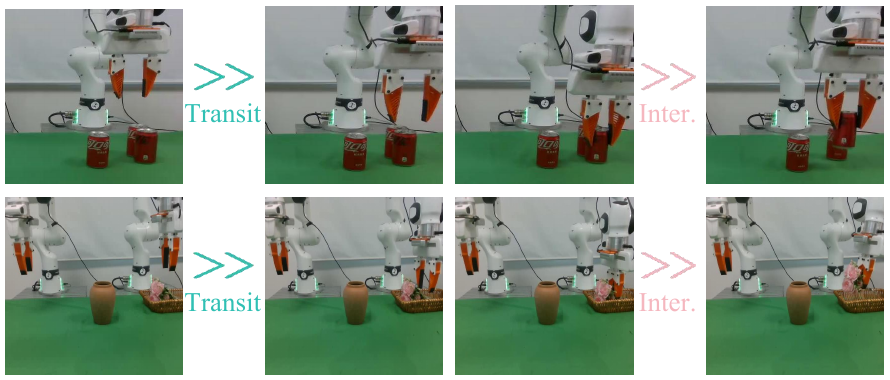}
  \end{minipage}
  \hfill
  \begin{minipage}[c]{0.55\linewidth}
    \centering
    \resizebox{\linewidth}{!}{
      \renewcommand{\arraystretch}{1.1} 
      \begin{tabular}{l c c c c c c} 
        \toprule
        \multirow{2}{*}{\textbf{Method}} & \multirow{2}{*}{\textbf{Domain}} & 
        \multirow{2}{*}{\textbf{Gen Scenarios}}
        & \multicolumn{2}{c}{\textbf{Stack Coke Cans}} & \multicolumn{2}{c}{\textbf{Put Flowers in Vase}} \\
        \cmidrule(lr){4-5} \cmidrule(lr){6-7}
        & & & Transit & Interaction & Transit & Interaction \\
        \midrule
        \multirow{4}{*}{\makecell[c]{\textbf{Imagine-}\\\textbf{then-}\\\textbf{Execute}}} & In & --- & 10/10 & 7/10 & 10/10 &  8/10  \\
        \cmidrule{2-7}
        & \multirow{3}{*}{OOD} & Background & 10/10 & 6/10  & 10/10 & 6/10 \\
        & & Position   & 10/10 & 5/10 & 10/10 & 2/10 \\
        & & Objects   & 10/10 & 7/10 & 10/10 & 7/10 \\
        \midrule
        \multirow{4}{*}{\makecell[c]{\textbf{Joint}\\\textbf{Modeling}}} & In & --- & 9/10 & 9/10 & 9/10 & 10/10 \\
        \cmidrule{2-7}
        & \multirow{3}{*}{OOD} & Background & 5/10 & 8/10$^*$ & 5/10 & 9/10$^*$ \\
        & & Position   & 3/10 & 10/10$^*$  & 0/10 & 10/10$^*$ \\
        & & Objects   & 0/10 & 10/10$^*$ & 6/10 & 10/10$^*$ \\
        \bottomrule
      \end{tabular}
    }
  \end{minipage}
  \vspace{-0.45cm}
\end{table}

%% file: sec/5_experiments.tex
\section{Experiments}
\label{headings}
Section~\ref{sec4.1} introduces the experimental setup. Section~\ref{sec4.2} reports results on real-world in-domain (ID) tasks, while Section~\ref{sec4.3} evaluates performance on out-of-domain (OOD) scenarios to assess generalization. Section~\ref{sec4.4} presents ablation studies of key components.

\subsection{Experiment Setup}
\label{sec4.1}

\textbf{Data Collection.} 
We evaluate HarmoWAM on real-world manipulation tasks using dual-arm Franka Research 3 robots, each equipped with three Intel RealSense cameras (one third-person and two wrist-mounted views).
We collect demonstrations for four single-arm tasks—\textit{Pick Fruit to Plate}, \textit{Stack Coke Cans}, \textit{Pour Coke into Beaker}, and \textit{Write ``Yes''}, and two dual-arm collaborative tasks, \textit{Put Flowers in Vase} and \textit{Put Items to Bag and Zip}.
Each task includes 100 demonstration trajectories collected via SpaceMouse teleoperation. Details are provided in Appendix~\ref{app:setup} and~\ref{app:data}.

\textbf{Training and Evaluation Details.}
We compare against five baselines spanning three categories: (1) SOTA VLA policies, including $\pi_{0.5}$~\cite{2025_4_22_pi0_5} and QwenVLA-OFT~\cite{bai2025qwen3}; (2) Imagine-then-Execute WAMs, instantiated using Wan2.2-TI2V-5B~\cite{wan2025wan} for future video generation and AnyPos~\cite{tan2025anypos} for action inference, following~\cite{chi2025wow}; and (3) Joint Modeling WAMs, including VPP~\cite{hu2025videopredictionpolicygeneralist} and Cosmos-Policy~\cite{kim2026cosmos}, with training details and additional descriptions provided in Appendix~\ref{app:baselines}.
For complex dual-arm tasks, we compare only the strongest baseline in each category, as others perform markedly worse.
Each method is evaluated over 20 independent episodes per task with randomized tabletop object positions.
We report the final success rate over long-horizon tasks (human evaluation). Due to space limitations, per-step (atomic) task performance is provided in Appendix~\ref{app:additional_results_in} and ~\ref{app:additional_results_out}.

\begin{figure}[t] 
    \centering
    \vspace{-0.1cm}
    \includegraphics[width=0.99\textwidth]{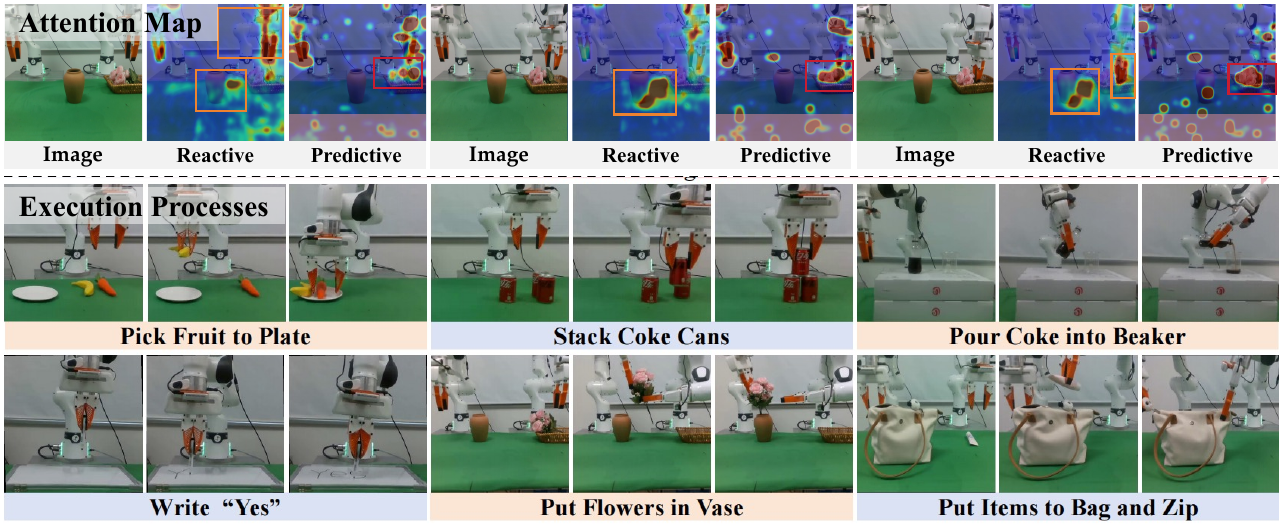} 
    \vspace{-0.2cm}
    \caption{The upper part presents attention map visualizations from the last-layer features of the reactive and predictive experts, while the lower part illustrates the robot execution process.}
    \vspace{-0.5cm}
    \label{fig:real_execution}
\end{figure}
\input{tables/experiments_in_all}

\subsection{In-Domain Results}
\label{sec4.2}
As shown in Table~\ref{tab:in_domain_all}, HarmoWAM achieves the best performance in ID real-world evaluations. HarmoWAM achieves 89\%, outperforming $\pi_{0.5}$ at 74\%, Cosmos-Policy at 78\% and Wan+AnyPos at 67\%. We additionally compute the standard deviation of our method ($\pm$ 3\%) by repeating the evaluation three times under the same setting. HarmoWAM demonstrates strong advantages on tasks requiring high precision and long horizons. Specifically, on \textit{Stack Coke Cans}, it achieves 90\% success, highlighting its ability to accurately align objects during stacking.
HarmoWAM attains 85\% success on the long-horizon task \textit{Put Items to Bag and Zip}, outperforming the next-best baseline by 13\%, indicating improved robustness in both transit and interaction phases. It also achieves 85\% on \textit{Put Flowers in Vase}, demonstrating effective bimanual coordination during handover and insertion.
These results suggest that our proposed framework effectively unlocks world model spatio-temporal knowledge, enabling coordinated control between two experts for robust manipulation across diverse tasks. In Figure~\ref{fig:real_execution}, we visualize real-world execution, with additional visualizations, failure cases, and videos provided in Appendix~\ref{app:Additional Task Execution Visualizations}, Appendix~\ref{app:failure}, and the supplementary material.

\subsection{Generalization}
\label{sec4.3}
To evaluate zero-shot generalization, we conduct experiments under multiple unseen configurations beyond the training data. We consider three OOD scenarios: unseen backgrounds, object positions, and manipulated objects. 
Representative examples are shown in Figure~\ref{fig:gene}, while the complete examples and detailed OOD settings are provided in Appendix~\ref{app:Additional Generalization Visualizations} and ~\ref{app:OOD_details}.

\textbf{Unseen Background.}
As shown in Table~\ref{tab:ood_all}, in cluttered OOD scenarios with unseen distractors and lighting changes, HarmoWAM maintains an 81\% success rate, outperforming Wan+AnyPos (53\%) and $\pi_{0.5}$ (60\%). 
Compared methods often suffer from visual misgrounding, confusing distractors with targets during transit.
In contrast, HarmoWAM leverages a world model to predict task-relevant visual evolution, providing generalizable semantic guidance for two action experts and enabling coordinated execution for reliable performance in visually complex OOD environments.
\input{tables/experiments_out_all}

\textbf{Unseen Position.}
To evaluate spatial generalization, we construct OOD regions at test time that lie outside the training data distribution. This represents the most challenging generalization scenario. In these regions, $\pi_{0.5}$ exhibits a significant drop to 32\%, and Cosmos-Policy to 26\%.
The Imagine-then-Execute paradigm improves generalizable transit via video prediction, but its inverse-dynamics-based action inference still lacks precision. For example, even in the simple \textit{Pick Fruit} task, the robot can reach the OOD region but fails to achieve accurate grasping.
In contrast, HarmoWAM mitigates this transit–interaction trade-off through world-model-guided expert collaboration, where the world model provides a shared spatio-temporal structure that decouples exploration and precision: it enables the reactive expert to generalize beyond the SFT distribution during transit, while allowing the predictive expert to exploit temporally coherent dynamics for precise interaction.

\begin{figure*}[t] 
    \centering
    \includegraphics[width=\textwidth]{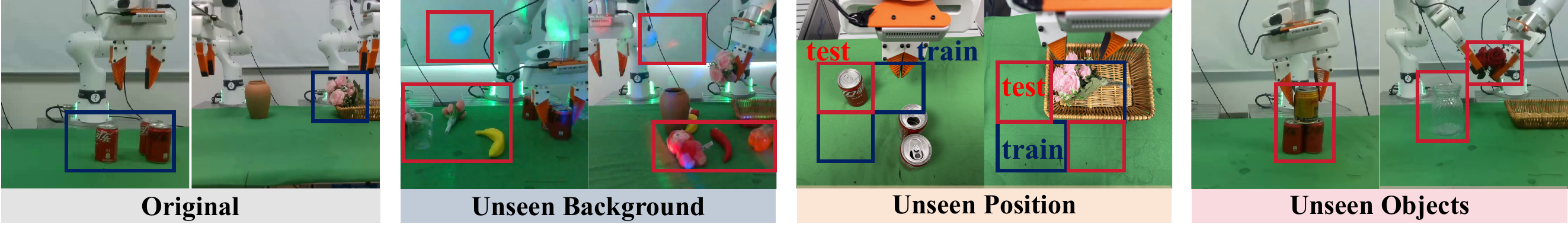} 
    \vspace{-0.4cm}
    \caption{\textbf{Generalization experiments.} Red boxes highlight unseen objects, background variations, and manipulated object positions, while blue boxes indicate original training configurations.}
    \vspace{-0.1cm}
    \label{fig:gene}
\end{figure*}

\textbf{Unseen Objects.}
We introduce unseen objects in each task to evaluate semantic generalization. All other methods exhibit significant performance degradation, whereas HarmoWAM achieves an 85\% zero-shot success rate.
This further demonstrates that our approach effectively leverages the semantic knowledge encoded in the world model, using both explicit and implicit conditioning to coordinate the two action experts for reliable object approach and interaction with unseen objects.

\subsection{Ablation Study}
\label{sec4.4}
To validate the effectiveness of each component, we conduct detailed ablation studies on the \textit{Put Flowers in Vase} and \textit{Pick Fruit to Plate} tasks, and report the average performance across them. These tasks cover long-horizon execution, bimanual coordination, and precise manipulation.

\begin{figure*}[t] 
    \centering
    \vspace{-0.2cm}
    \includegraphics[width=\textwidth]{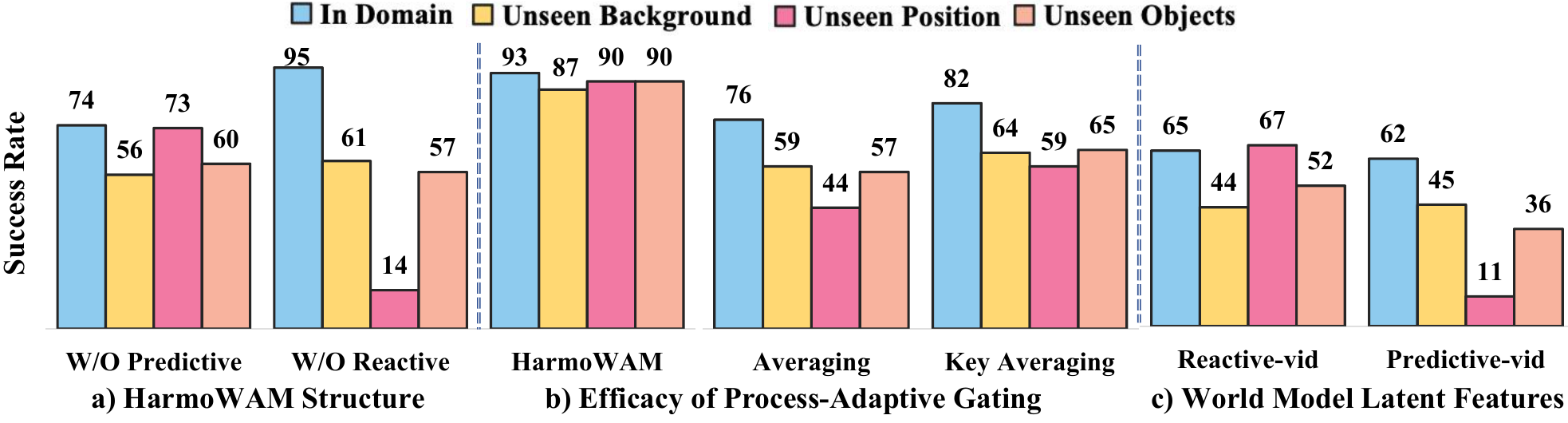} 
    \caption{\textbf{Ablation Study.} We investigate (a) HarmoWAM Structure, (b) Efficacy of Process-Adaptive Gating, and (c) Impact of world model latent features on both action experts. The ``-vid” suffix indicates that video latent features are excluded from the action expert’s conditioning.}
    \vspace{-0.5cm}
    \label{fig:ablation}
    
\end{figure*}

\textbf{HarmoWAM Structure.}
As shown in Figure~\ref{fig:ablation} a), we first validate the effectiveness of the HarmoWAM architecture and the necessity of each expert. Removing the reactive expert causes the success rate under position OOD to drop to 14\%, indicating that the predictive expert alone struggles to effectively translate the world model knowledge into actionable signals for expanding the exploration space.
Conversely, removing the predictive expert reduces the success rates under position OOD and object-instance OOD to 56\% and 60\%, respectively. 
This suggests that the reactive expert, which relies on predicted visual observations, is insufficient for fine-grained action generation under visual and object variations.
These results reveal that the two experts play complementary roles under world model guidance: exploration emerges from reactive inference, while precision arises from predictive reasoning, making their coordinated interaction essential rather than a simple aggregation.

\textbf{Efficacy of Process-Adaptive Gating.}
We compare Process-Adaptive Gating with two alternative routing strategies: (1) Averaging, which numerically averages the outputs of two experts at each timestep; and (2) Keyframe-Based Averaging, which applies averaging during the interaction phase when the predictive expert is activated.
All variants exhibit performance degradation under OOD settings. For example, in the position OOD scenario, Averaging leads to a 46\% drop in performance, while Keyframe-Based Averaging reduces the drop to 31\%.
In contrast, Process-Adaptive Gating performs stage-aware coordination based on real-time visual feedback. This reveals that effective manipulation relies on stage-aware use of world model knowledge, enabled by the gating mechanism, which routes control between complementary experts for generalizable transit and precise interaction.

\textbf{World Model Latent Features.}
World model’s video latent features play a crucial role in conditioning both action experts. Removing these features from the reactive expert reduces its ID performance to 65\% and its OOD average to 54\%. Similarly, removing them from the predictive expert causes its ID performance to drop from 95\% to 62\%. These results demonstrate that the world model’s latent features enhance temporally coherent understanding and physical world modeling in both experts.
The additional ablation study on the denoising steps of the world model is provided in Appendix~\ref{app:ablation}.

%% file: tables/experiments_in_all.tex
\begin{table}[t]
  \centering
  \small
  \caption{\textbf{In-Domain Quantitative Results on Real-World Tasks.} Task-level success rates for four single-arm and two dual-arm tasks, computed as the average over all critical sub-stages.}
  \label{tab:in_domain_all}

  \setlength{\aboverulesep}{0pt}
  \setlength{\belowrulesep}{0pt}
  \renewcommand{\arraystretch}{1.2} 

  \resizebox{\textwidth}{!}{
  \begin{tabular}{l c c c c c c c}
    \hline
    \multirow{2}{*}{\textbf{Method}} & \multicolumn{4}{c}{\textbf{Single-arm Tasks (Avg)}} & \multicolumn{2}{c}{\textbf{Dual-arm Tasks (Avg)}} & \multirow{2}{*}{\textbf{Avg}} \\
    \cmidrule(lr){2-5} \cmidrule(lr){6-7}
    & \textbf{Pick Fruit} & \textbf{Stack Cans} & \textbf{Pour Coke} & \textbf{Write "Yes"} & \textbf{Put Flowers} & \textbf{Put Items} & \\
    \hline

    $\pi_{0.5}$     & 0.80 & 0.68 & 0.75 & 0.83 & 0.72 & 0.67 & 0.74 \\
    VPP             & 0.80 & 0.60 & 0.78 & 0.73 & ---  & ---  & 0.73 \\
    Wan+Anypos      & 0.88 & 0.60 & 0.78 & 0.72 & 0.53 & 0.52 & 0.67 \\
    QwenVLA-OFT     & 0.78 & 0.30 & 0.73 & 0.72 & ---  & ---  & 0.63 \\
    Cosmos-Policy   & 0.93 & 0.65 & 0.80 & 0.83 & 0.75 & 0.72 & 0.78 \\
    \midrule
    \rowcolor{blue!8} \textbf{Ours} & \textbf{0.95} & \textbf{0.90} & \textbf{0.88} & \textbf{0.92} & \textbf{0.85} & \textbf{0.85} & \textbf{0.89} \\
    \hline
  \end{tabular}
  } 
\vspace{-0.4cm}
\end{table}

%% file: tables/experiments_out_all.tex
\begin{table}[t]
  \centering
  \caption{\textbf{Generalization Performance.} Global Avg represents the average over all OOD settings.
  The percentage drop (in parentheses) is relative to the ID score.
  }
  \label{tab:ood_all}

  \setlength{\aboverulesep}{0pt}
  \setlength{\belowrulesep}{0pt}
  \renewcommand{\arraystretch}{1.2} 

  \resizebox{\textwidth}{!}{
  \begin{tabular}{l l c c c c c c c c}
    \hline
    \multirow{2}{*}{\textbf{Method}} & \multirow{2}{*}{\textbf{Gen Scenario}} & \multicolumn{4}{c}{\textbf{Single-arm Tasks (Avg)}} & \multicolumn{2}{c}{\textbf{Dual-arm Tasks (Avg)}} & \multirow{2}{*}{\textbf{Avg}} & \textbf{Global} \\
    \cmidrule(lr){3-6} \cmidrule(lr){7-8}
    & & \textbf{Pick Fruit} & \textbf{Stack Cans} & \textbf{Pour Coke} & \textbf{Write "Yes"} & \textbf{Put Flowers} & \textbf{Put Items} & & \textbf{Avg} \\
    \hline

    \multirow{3}{*}{$\pi_{0.5}$}
    & Background & 0.65 & 0.40 & 0.63 & 0.67 & 0.62 & 0.60 & 0.60 & \multirow{3}{*}{%
  \begin{tabular}{@{}c@{}}
    0.49\\[-1pt]
    {\small\textcolor{red}{($33.8\%\,\downarrow$)}}
  \end{tabular}%
} \\
    & Position   & 0.20 & 0.08 & 0.43 & 0.55 & 0.37 & 0.26 & 0.32 & \\
    & Objects   & 0.55 & 0.43 & 0.53 & 0.63 & 0.60 & 0.49 & 0.54 & \\
    \hline

    \multirow{3}{*}{VPP}
    & Background & 0.38 & 0.25 & 0.68 & 0.40 & ---  & ---  & 0.43 & \multirow{3}{*}{%
  \begin{tabular}{@{}c@{}}
    0.41\\[-1pt]
    {\small\textcolor{red}{($43.8\%\,\downarrow$)}}
  \end{tabular}%
} \\
    & Position   & 0.28 & 0.10 & 0.20 & 0.33 & ---  & ---  & 0.23 & \\
    & Objects   & 0.68 & 0.40 & 0.68 & 0.50 & ---  & ---  & 0.57 & \\
    \hline

    \multirow{3}{*}{Wan+Anypos}
    & Background & 0.58 & 0.43 & 0.65 & 0.65 & 0.47 & 0.38 & 0.53 & \multirow{3}{*}{%
  \begin{tabular}{@{}c@{}}
    0.53\\[-1pt]
    {\small\textcolor{red}{($20.9\%\,\downarrow$)}}
  \end{tabular}%
} \\
    & Position   & 0.83 & 0.40 & 0.33 & 0.50 & 0.45 & 0.45 & 0.49 & \\
    & Objects   & 0.63 & 0.65 & 0.68 & 0.68 & 0.40 & 0.43 & 0.58 & \\
    \hline

    \multirow{3}{*}{QwenVLA-OFT}
    & Background & 0.48 & 0.13 & 0.58 & 0.63 & ---  & ---  & 0.46 & \multirow{3}{*}{%
  \begin{tabular}{@{}c@{}}
    0.41\\[-1pt]
    {\small\textcolor{red}{($34.9\%\,\downarrow$)}}
  \end{tabular}%
} \\
    & Position   & 0.28 & 0.10 & 0.33 & 0.42 & ---  & ---  & 0.28 & \\
    & Objects   & 0.63 & 0.10 & 0.73 & 0.52 & ---  & ---  & 0.50 & \\
    \hline
    
    \multirow{3}{*}{Cosmos-Policy}
    & Background & 0.85 & 0.38 & 0.73 & 0.57 & 0.42 & 0.48 & 0.57 & \multirow{3}{*}{%
  \begin{tabular}{@{}c@{}}
    0.44\\[-1pt]
    {\small\textcolor{red}{($43.6\%\,\downarrow$)}}
  \end{tabular}%
} \\
    & Position   & 0.20 & 0.20 & 0.15 & 0.55 & 0.25 & 0.21 & 0.26 & \\
    & Objects   & 0.80 & 0.38 & 0.68 & 0.22 & 0.50 & 0.44 & 0.50 & \\
    \hline

    \rowcolor{blue!8} & Background & \textbf{0.90} & \textbf{0.78} & \textbf{0.83} & \textbf{0.82} & \textbf{0.73} & \textbf{0.78} & \textbf{0.81} & \\
    \rowcolor{blue!8} & Position   & \textbf{0.93} & \textbf{0.78} & \textbf{0.80} & \textbf{0.83} & \textbf{0.80} & \textbf{0.64} & \textbf{0.80} & \\
    \rowcolor{blue!8} \multirow{-3}{*}{Ours} & Objects   & \textbf{0.88} & \textbf{0.85} & \textbf{0.88} & \textbf{0.87} & \textbf{0.80} & \textbf{0.81} & \textbf{0.85} & \multirow{-3}{*}{%
  \begin{tabular}{@{}c@{}}
    \textbf{0.82}\\[-1pt]
    {\small\textcolor{red}{($7.9\%\,\downarrow$)}}
  \end{tabular}%
}\\
    \hhline{----------} 
  \end{tabular}
  }
  \vspace{-0.2cm}
\end{table}

%% file: sec/7_appendix.tex
\section{Real-World Set-up}

\label{app:setup}

\begin{figure*}[htbp]
    \centering
    \includegraphics[width=\textwidth]{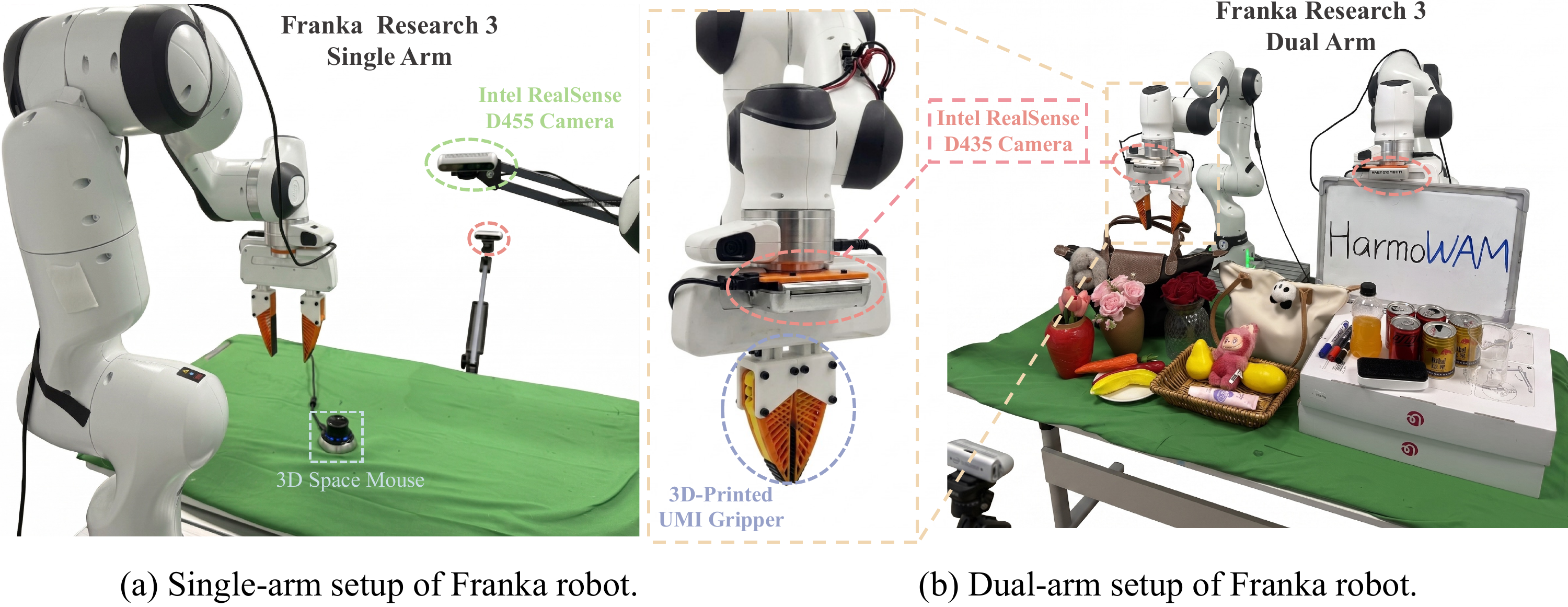}
    \caption{Real-World robot setup and experimental assets.}
    \label{fig:Robot}
\end{figure*}

\textbf{Single-Arm Configuration.} 
As shown in Figure~\ref{fig:Robot}, our single-arm platform is built on a 7-DoF Franka Research 3 (FR3) manipulator equipped with a 3D-printed UMI gripper for contact-rich precision manipulation tasks. The perception system employs three Intel RealSense cameras to provide multi-view visual feedback: a stationary front-view D435 positioned at approximately $45^\circ$ above the workspace captures a global view, a top-down D455 mounted directly above the manipulation surface provides precise spatial localization, and a wrist-mounted D435 attached to the end-effector supplies close-range observations. All cameras operate at $640 \times 480$ resolution. At each timestep, the policy observation consists of three RGB images and the robot's proprioceptive state.

\textbf{Dual-Arm Configuration.} 
The dual-arm platform extends to two parallel FR3 arms, each equipped with the same UMI gripper. The camera layout includes one stationary global camera and one wrist-mounted camera per arm. Thus, each dual-arm observation also contains three RGB images, but with two wrist views and one global view. The proprioceptive state is extended to 14 dimensions, with 7 dimensions per arm.

\textbf{Control and Data Collection.} 
The end-effector poses, gripper states, and other proprioceptive signals of each arm are synchronized with camera observations through a real-time communication interface. All demonstrations are collected via teleoperation using a SpaceMouse 3D controller, with the operator receiving real-time feedback from all camera views.

\section{Self-Collected Data}
\label{app:data}

\textbf{Single-Arm Tasks}

\textit{1. Pick Fruit to Plate.} The robot sequentially picks up a banana and a carrot and places them onto a plate. $S_1$: grasp and place banana; $S_2$: grasp and place carrot. The average length is approximately 280 steps and is sequentially evaluated. A stage is considered successful if the robot successfully grasps the corresponding object and places it stably on the plate without dropping it.

\textit{2. Stack Coke Cans.} The robot stacks three cans one by one, demanding highly precise spatial alignment. $S_1$: place the second can beside the first; $S_2$: place the third can on top. The average length is approximately 290 steps and is sequentially evaluated. $S_1$ is considered successful if the robot places the second can beside the first can with stable contact and without knocking over or displacing the first can. 
$S_2$ is considered successful if the robot places the third can on top of the existing cans with stable stacking and without collapsing the stack.

\textit{3. Pour Coke into Beaker.} The robot grasps a bottle and pours its contents into a beaker, testing fine-grained rotational control. $S_1$: grasp bottle; $S_2$: pour into beaker. The average length is approximately 310 steps and is sequentially evaluated. $S_1$ is considered successful if the robot securely grasps and lifts the bottle, and $S_2$ is considered successful if the robot tilts the bottle toward the beaker and pours the contents into the beaker without spilling outside the target region.

\textit{4. Write ``Yes''.} The robot picks up a marker and writes ``Y'', ``e'', ``s'' on a whiteboard in sequence. $S_1$: write ``Y''; $S_2$: write ``e''; $S_3$: write ``s''. The average length is approximately 310 steps and is sequentially evaluated. A stage is considered successful if the robot writes the corresponding character legibly on the whiteboard.

\textbf{Dual-Arm Tasks}

\textit{1. Put Flowers in Vase.} The left arm picks a flower and hands it to the right arm, which inserts it into a vase, requiring precise bimanual coordination and tight-tolerance insertion. $S_1$: pick flower; $S_2$: bimanual handover; $S_3$: insert into vase. The average length is approximately 280 steps and is sequentially evaluated. $S_1$ is considered successful if the left arm securely grasps and lifts the flower without dropping it. $S_2$ is considered successful if the flower is successfully transferred from the left arm to the right arm while maintaining a stable grasp and without dropping the flower. $S_3$ is considered successful if the right arm inserts the flower into the vase.

\textit{2. Put Items to Bag and Zip.} Both arms collaborate to place items into a bag and zip it closed, which is the longest-horizon task. $S_1 \to S_2$: pick up item and place into bag; $S_3 \to S_4 \to S_5$: one arm grips the bag to hold it steady, the other grips and pulls the zipper to close. The average length is approximately 400 steps and is sequentially evaluated. $S_1$ is considered successful if the robot securely grasps and lifts the target item without dropping it. $S_2$ is considered successful if the item is placed inside the bag. $S_3$ is considered successful if one arm firmly holds the bag in place without significantly deforming or displacing it. $S_4$ is considered successful if the other arm successfully grasps the zipper puller. $S_5$ is considered successful if the zipper is pulled along the closing direction and the bag is closed.

\section{Real-World Experiment Details.}
\label{app:details}

\subsection{Baselines.}
\label{app:baselines}
\textbf{$\pi_{0.5}$:}~\cite{2025_4_22_pi0_5} A PaliGemma-based VLA with a hierarchical architecture that generates continuous actions via flow matching.
\textbf{QwenVLA-OFT:}~\cite{bai2025qwen3} An implementation that adopts Qwen3-VL as the backbone and performs parallel action prediction with a $\ell_1$ regression objective.
\textbf{VPP:}~\cite{hu2025videopredictionpolicygeneralist} A video-prediction-based policy that extracts intermediate features from a video diffusion model to condition a separate diffusion policy head.
\textbf{Cosmos-Policy:}~\cite{kim2026cosmos} A joint video-action modeling baseline that encodes actions as latent-space frames and denoises them together with video frames in the Cosmos video model.
\textbf{Wan+AnyPos:}~\cite{wan2025wan,tan2025anypos} An Imagine-then-Execute baseline where Wan2.2-TI2V-5B predicts future trajectory videos and AnyPos serves as the Inverse Dynamics Model to extract actions from video frame pairs.

For all baselines, we follow the training protocols recommended in their original papers whenever applicable. 
$\pi_{0.5}$, Cosmos-Policy, and Wan2.2-TI2V-5B are initialized from their officially released pretrained checkpoints and then fine-tuned using the configurations and hyperparameters suggested by the corresponding original papers. 
For AnyPos, we train the model from scratch on the downstream dataset following the recommended settings in the original paper. For VPP, we fine-tune Stable Video Diffusion and replace its action head with the same pretrained diffusion head used in our method to ensure a fair comparison. QwenVLA-OFT is built upon Qwen3-VL-4B. Following prior work~\cite{liu2026last}, we additionally pretrain the model on over 400K cross-embodiment trajectories to endow it with robotic priors, and further fine-tune it for robotic control using a parallelized $\ell_1$ regression objective.
All methods are trained on the same real-world demonstration dataset, given the same task instructions, and evaluated using the same multi-view RGB observations, except for Wan+AnyPos, which uses a third-view camera perspective. All methods predict actions in the same control space: 7 dimensions for single-arm tasks and 14 dimensions for dual-arm tasks. We evaluate all methods under the same real-world testing protocol.

\input{tables/pretrain}
\subsection{Training Protocol.}
\label{app:trainingdetails}
All models are trained on 8 NVIDIA H20 GPUs. World model is initialized from a Wan2.2-TI2V-5B~\cite{wan2025wan} backbone pretrained on broad robot datasets~\cite{lightewm2026}, including publicly available datasets such as DROID~\cite{khazatsky2024droid}, AgiBot~\cite{bu2025agibot}, and RoboMIND~\cite{wu2024robomind}, together with closed-source robot data. Detailed statistics of the public datasets are provided in Table~\ref{tab:dataset_statistics}. We then perform full fine-tuning on our real-world demonstration dataset for task-specific video prediction using a Flow Matching objective. In the second stage, the world model is frozen, while the predictive action expert, the reactive action expert, and the Process-Adaptive Gating network are optimized using the explicit and implicit video conditions produced by the world model. The predictive expert is trained with a diffusion denoising loss, the reactive expert with a Smooth L1 loss, and the gating network with a binary cross-entropy loss. Specifically, the loss of the reactive expert is defined as:
\begin{equation}
\mathcal{L}_{\mathrm{react}} = \mathbb{E} \left[ d(\hat{\mathbf{a}}_{t+1:t+H}, \mathbf{a}_{t+1:t+H}) \right].
\end{equation}
where \(d(\cdot,\cdot)\) denotes the Smooth L1 distance:
\begin{equation}
d(x,\hat{x}) =
\begin{cases}
0.5 \cdot \frac{(x-\hat{x})^2}{\beta}, & \text{if } |x-\hat{x}| < \beta, \\
|x-\hat{x}| - 0.5\beta, & \text{otherwise},
\end{cases}
\end{equation}
where $\beta$ is set to $0.1$. The overall objective is a weighted sum of all subs-losses. The action space is 7-dimensional for single-arm tasks and 14-dimensional for dual-arm tasks.

\subsection{Out-of-Domain (OOD) Scenario Construction}
\label{app:OOD_details}

\textbf{Unseen Background.}
To evaluate robustness against visual context shifts, we construct background OOD scenes by introducing unseen distractor objects and lighting changes. Specifically, 5--8 objects that never appear during training are randomly placed in the workspace, with variations in color, texture, and geometry. These distractors are distributed near the target object and in peripheral tabletop regions to increase scene clutter. We further modify ambient lighting by changing light intensity and incident direction, producing appearance changes, shadows, and local reflections not observed during training. This setting tests the model's ability to maintain reliable perception and stable manipulation in cluttered and visually perturbed environments.

\textbf{Unseen Position.}
Target objects are placed in workspace regions outside the spatial coverage of training demonstration trajectories. Specifically, we partition the manipulation surface into training and test regions, where the test region is spatially disjoint from the regions covered by object positions during data collection. This setting evaluates the model's generalization capability under unseen spatial coordinates.

\textbf{Unseen Objects.}
Target objects in each task are replaced with substitutes that differ significantly from the training objects in visual appearance and geometry. For example, a carrot is replaced with a pepper, a standard Coke can is replaced with a Red Bull can with different body dimensions and shape, and the original flower is replaced with flowers of different colors and varying numbers of blossoms. This setting evaluates the model's robustness to object-level semantic and geometric variations, particularly when grasping strategies and interaction patterns need to be adaptively adjusted.

\section{Additional Method Details}

\subsection{Additional Preliminaries}
\label{app:preli}
\textbf{IDM-based Action Inference.} The world model predicts future visual frames and spatio-temporal latent representations, which are not directly executable by the robot. To translate these predictions into low-level control actions, some works~\cite{tan2025anypos,mi2026tc, feng2025vidar} adopt an Inverse Dynamics Model (IDM), denoted by $\pi_{\mathrm{IDM}}$. Given a future video trajectory $\mathbf{V}_{t:t+H}$, the IDM produces the action sequence:
\(
\mathbf{a}_{t+1:t+H} \sim \pi_{\mathrm{IDM}}(\cdot \mid \mathbf{V}_{t+1:t+H}).
\)
The output follows the same action paradigm as defined above.

\begin{figure*}[htbp]
    \centering
    \includegraphics[width=\textwidth]{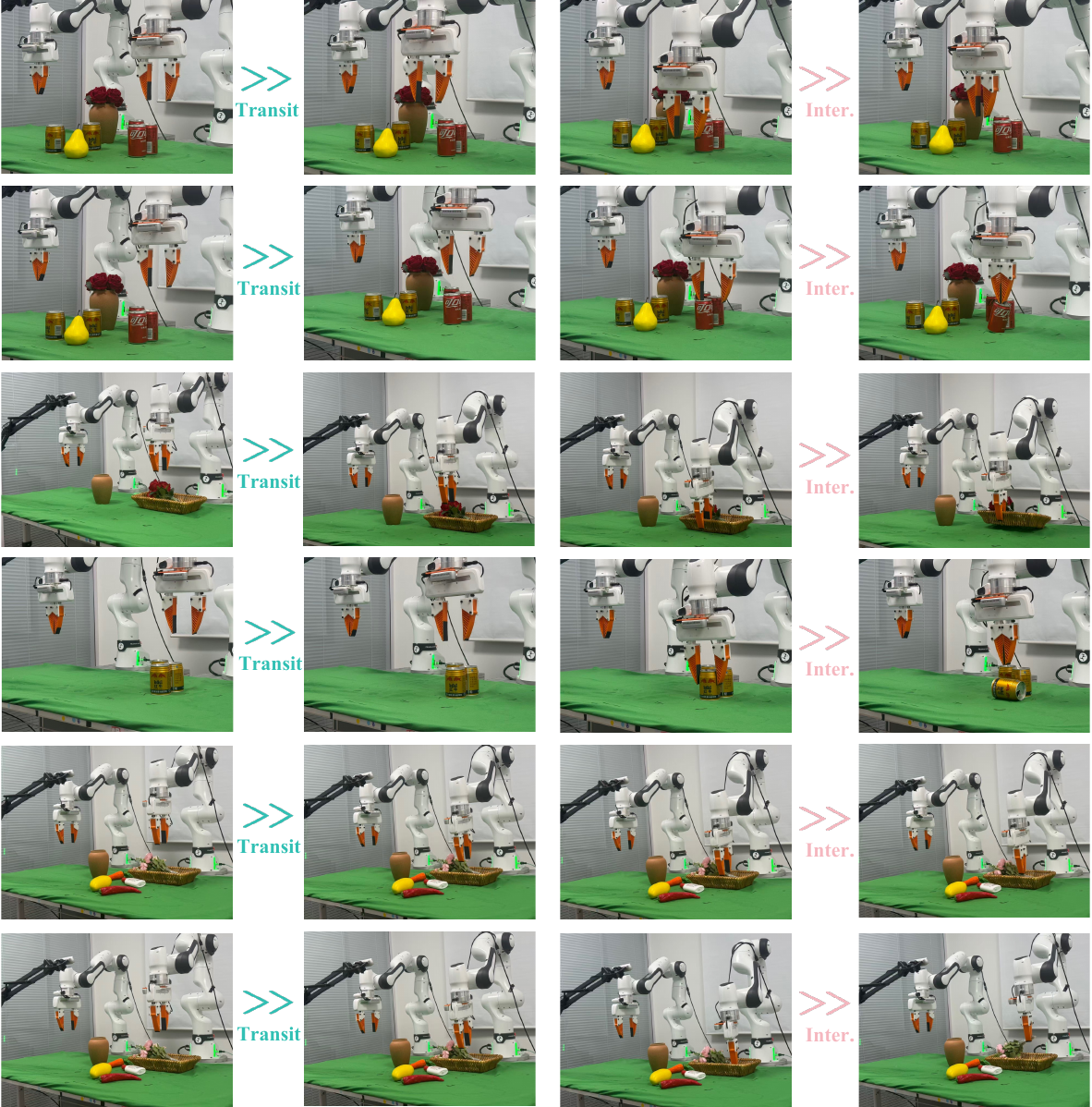}
    \caption{Representative failure cases of the two World Action Models paradigms under OOD scenarios. The first three rows show typical failures of the \emph{Joint Modeling} baseline and the last three rows show common failures of the \emph{Imagine-then-Execute} baseline. }
    \label{fig:motivation_aba}
\end{figure*}

\begin{figure}[htbp] 
    \centering
    \includegraphics[width=\textwidth]{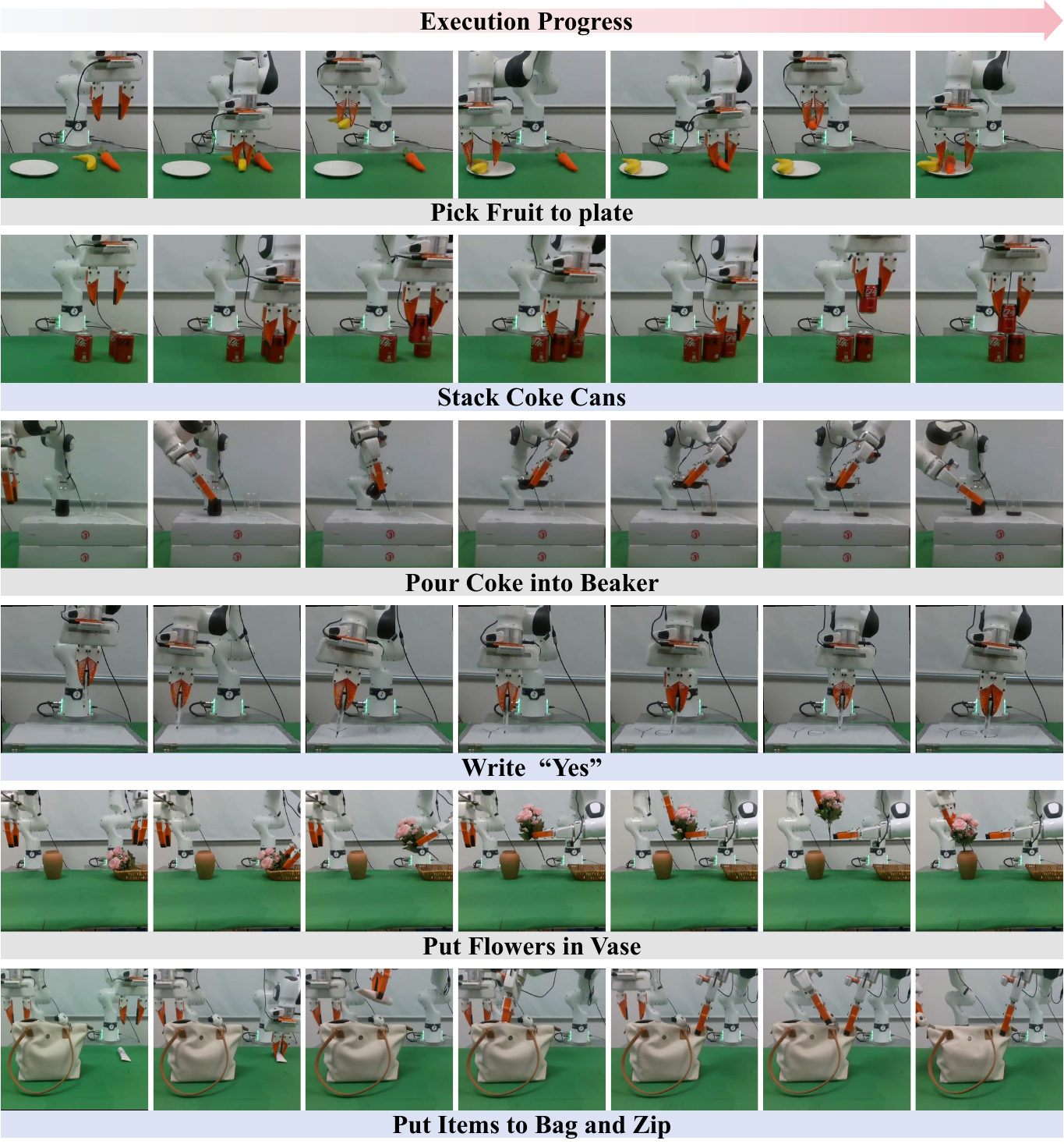} 
    \caption{Visualization of complete execution sequences on six real-world manipulation tasks.}
    \label{fig:additional_real}
    \vspace{-0.1cm}
\end{figure}

\subsection{Motivation Experiments Details}
\label{app:motivation}
We conduct the motivation study on two representative real-world tasks, \textit{Put Flowers in Vase} and \textit{Stack Coke Cans}, under both in-domain (ID) settings and three out-of-domain (OOD) scenarios, including background, position, and semantic variations, with detailed OOD settings provided in Appendix~\ref{app:OOD_details} and OOD visualizations shown in Appendix~\ref{app:Additional Generalization Visualizations}. For each task and domain setting, we evaluate both the \textit{Transit} and \textit{Interaction} phases over 10 real-world trials. \textit{Transit} measures the ability to reach the target vicinity from the initial position or move between objects, reflecting generalization. \textit{Interaction} measures the success of core actions such as picking, placing, stacking, handover, or insertion, reflecting precision.
For the \emph{Joint Modeling} baseline, the interaction performance is evaluated by initializing the robot near the target object. This design decouples interaction precision from transit failures, allowing us to evaluate manipulation capability independently of target-reaching success.
Figure~\ref{fig:motivation_aba} visualizes representative failure cases of the two paradigms. The first three rows show typical failures of the \emph{Joint Modeling} baseline. Under OOD scenarios, changes in background appearance or object position can shift the predicted trajectory away from the target region. As a result, the end-effector fails to move above the manipulated object or stops at an incorrect spatial location, causing failure in the transit phase before meaningful object contact is established. The last three rows show common failures of the \emph{Imagine-then-Execute} baseline. Although it can often generate a reasonable transit trajectory and move the gripper to the vicinity of the target object, the inferred actions may still be inaccurate during the interaction phase. For example, the gripper may close at an offset grasp point, or open or close prematurely, leading to unsuccessful grasping, placement, or stacking.

\subsection{Process-Adaptive Gating Details}
\label{app:gating_details}
The Process-Adaptive Gating Mechanism determines when HarmoWAM switches between the predictive action expert and the reactive action expert during execution. 
To train the gating module, we automatically construct frame-level supervision labels from demonstrations through a keyframe annotation pipeline. The pipeline identifies key interaction events from robot proprioceptive signals~\cite{shridhar2023perceiver}, including gripper state changes and task-specific end-effector height thresholds. 
Specifically, gripper transitions between two consecutive frames, including open-to-close transitions for grasping and close-to-open transitions for releasing, are marked as key events. 
In addition, for tasks where the interaction stage is better characterized by vertical motion, such as insertion, pouring, or placing, the end-effector height threshold is used as an auxiliary cue. 
These events usually correspond to central moments of interaction phases that require high-precision manipulation.

Based on these key events, the pipeline automatically labels a temporal window of 20 frames before and after each key frame as an interaction segment, since neighboring frames typically involve adjusting the pose, establishing contact, or completing the release.
Frames outside these interaction windows are labeled as transit segments, which mainly correspond to target approaching or coarse positioning across different manipulation regions.
For dual-arm tasks, if either arm satisfies the interaction-event criterion, the corresponding timestep and its surrounding temporal window are labeled as an interaction segment.
We further evaluate the gating module offline on held-out test demonstrations outside the training set. The gating classifier achieves an average frame-level accuracy of 96.95\%  1,637 test frame pairs, demonstrating its ability to reliably distinguish transit and interaction stages and effectively support adaptive expert switching during execution.

\section{Additional Visualization}

\subsection{Additional Task Execution Visualizations.} 
\label{app:Additional Task Execution Visualizations}
Figure~\ref{fig:additional_real} illustrates the complete execution sequences of HarmoWAM across all six tasks. We observe that HarmoWAM produces smooth and continuous motion trajectories, particularly in phases requiring fine-grained control, such as spatial alignment during can stacking, sustained angular control during pouring, and tight-tolerance alignment during flower insertion.

\subsection{Additional Generalization Visualizations.}
\label{app:Additional Generalization Visualizations}
\begin{figure*}[t] 
    \centering
    \includegraphics[width=\textwidth]{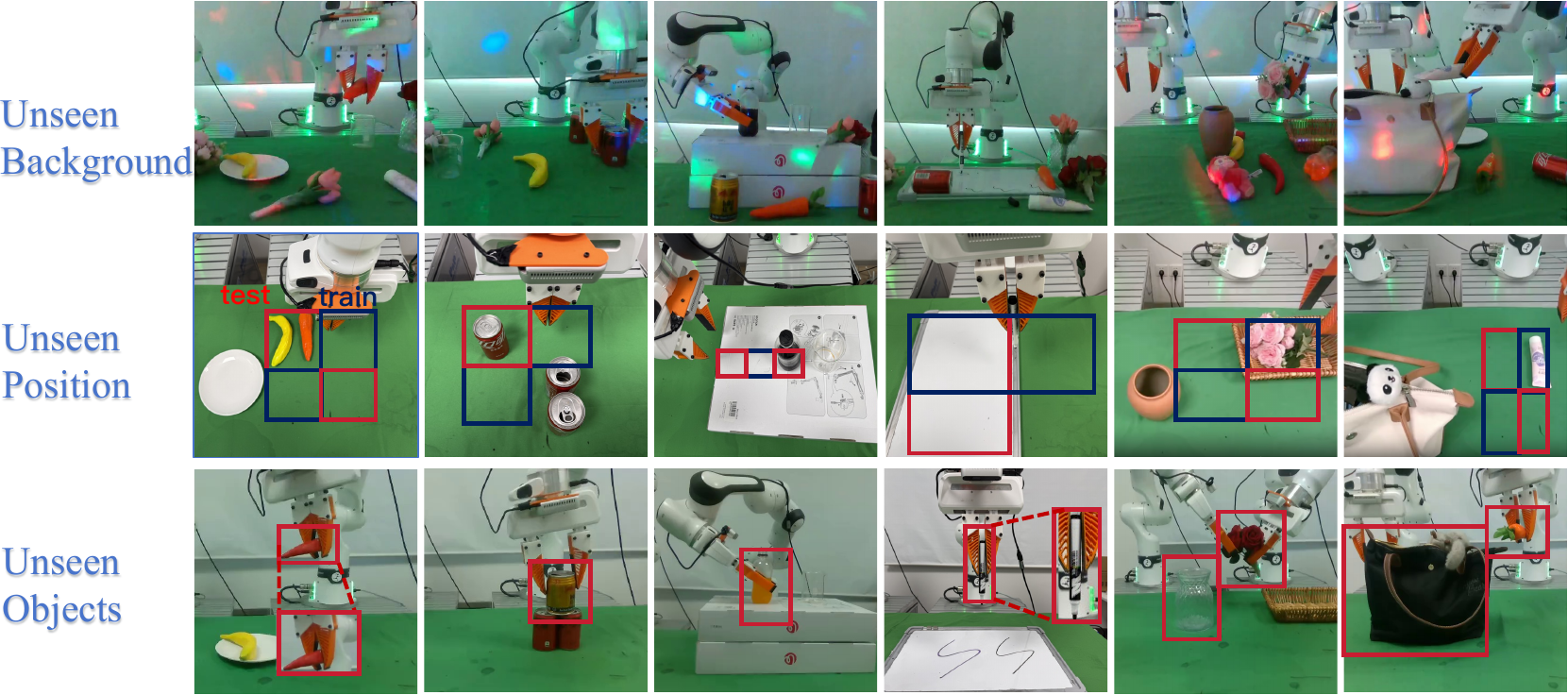} 
    \caption{Visualization of the complete examples under three OOD settings: unseen background, unseen position, and unseen objects.}
    \label{fig:gene_additional}
    \vspace{-0.1cm}
\end{figure*}

Figure~\ref{fig:gene_additional} shows the complete examples under all three OOD dimensions: background, position, and semantic variations. Scenario construction details are provided in Appendix~\ref{app:OOD_details}.

\section{Additional Quantitative and Qualitative Results}
\label{app:additional_results}

\input{tables/experiments_in}
\input{tables/experiments_in_dual}

\subsection{In-Domain(ID) Stage-Wise Results.}
\label{app:additional_results_in}
Tables~\ref{tab:in_domain} and~\ref{tab:in_domain_dual} report the complete stage-wise success rates on in-domain (ID) tasks, providing a fine-grained view of subgoal performance beyond task-level averages.
For single-arm tasks, HarmoWAM consistently achieves high success across different subgoals. In \textit{Stack Coke Cans}, which requires precise spatial alignment, HarmoWAM obtains 95\% success in the first stage and 85\% in the second stage, demonstrating reliable end-effector control across repeated stacking operations. In \textit{Pour Coke into Beaker}, which emphasizes continuous pose control, HarmoWAM achieves 90\% in the grasping stage and 85\% in the pouring stage, indicating that it can effectively maintain the relative position and orientation between the bottle and the beaker. Across all single-arm tasks, HarmoWAM reaches an average success rate of 91\%, outperforming the next-best baseline, Cosmos-Policy, by 10 percentage points.
For dual-arm tasks, the stage-wise results further highlight HarmoWAM's advantage in complex collaborative procedures. In \textit{Put Flowers in Vase}, most baselines can complete early picking or partial handover, but struggle with the final insertion stage. For example, Wan+AnyPos drops from 60\% in the handover stage to 15\% in the insertion stage, while HarmoWAM maintains 70\% success in the final stage. This shows that HarmoWAM better preserves precise control after bimanual handover. In \textit{Put Items to Bag and Zip}, which involves multiple consecutive subgoals such as item placement, bag stabilization, zipper grasping, and zipper pulling, baseline performance generally decreases in later stages. In contrast, HarmoWAM maintains 85\%, 80\%, and 80\% success across the final three stages, indicating stronger action coherence over long horizons and reduced error propagation. Overall, HarmoWAM achieves an average success rate of 85\% on dual-arm tasks, outperforming the next-best baseline, Cosmos-Policy, by 12 percentage points.
Overall, the ID stage-wise results further show that the advantage of HarmoWAM is not limited to higher task-level averages. It is also reflected in key subgoals such as stacking alignment, continuous pouring, insertion after bimanual handover, and zipper closure. These steps require finer local control than ordinary pick-and-place operations and are where baseline methods most frequently fail. 
These results highlight HarmoWAM’s advantage in using a world model to provide spatio-temporal conditions while the Process-Adaptive Gating coordinates the predictive action expert and the reactive action expert, enabling the framework to achieve reliable and accurate execution across complex tasks.

\subsection{Out-of-Domain(OOD) Stage-Wise Results.}
\label{app:additional_results_out}
\input{tables/experiments_out}
\input{tables/experiments_out_dual}
Tables~\ref{tab:ood} and~\ref{tab:ood_dual} present the OOD stage-wise success rates. The results show that different distribution shifts challenge different aspects of robot execution: background variation mainly affects visual grounding under distractors and illumination changes, position variation tests whether policies can act beyond the spatial distribution of the supervised training data, and semantic variation stresses the model's understanding of object geometry, functionality, and contact relations.
Under background variation, HarmoWAM achieves an average success rate of 83\% on single-arm tasks and 76\% on dual-arm tasks, outperforming the strongest baselines by 21 and 15 percentage points, respectively. Several baselines show clear degradation in grasping or placement stages under cluttered scenes. For example, in \textit{Stack Coke Cans}, the second-stage success of all baselines remains no higher than 35\%, while HarmoWAM reaches 65\%. In dual-arm tasks, HarmoWAM also maintains 60\% success in the final insertion stage of \textit{Put Flowers in Vase}, compared with 30\% for the best baseline. These results indicate that HarmoWAM remains robust by using the world model's predictive and generalization capabilities to forecast task-relevant visual dynamics while suppressing distractors, enabling downstream action pathways to focus on object interactions and reliably execute actions in cluttered environments.
Under position variation, the performance gap becomes more pronounced. HarmoWAM achieves 83\% average success on single-arm tasks and 70\% on dual-arm tasks, while the best baselines only reach 51\% and 45\%, respectively. This suggests that baseline models are strongly constrained by the spatial coverage of the supervised fine-tuning data. In \textit{Stack Coke Cans}, HarmoWAM achieves 85\% and 70\% across the two stages, whereas the best baseline drops to 55\% and 25\%. In \textit{Put Flowers in Vase}, HarmoWAM maintains 90\%, 85\%, and 65\% across the three stages, substantially outperforming Wan+AnyPos, which drops from 70\% to 15\% by the final insertion stage. These results show that HarmoWAM leverages the world model's generalization capability to guide the reactive action expert beyond the constraints of the SFT data distribution, while relying on the predictive action expert for fine-grained manipulation at critical interaction stages.
Under semantic variation, HarmoWAM achieves the highest average success rates, reaching 87\% on single-arm tasks and 81\% on dual-arm tasks, compared with the strongest baselines at 66\% and 53\%. The advantage is especially clear in contact-sensitive stages. For example, in \textit{Stack Coke Cans}, HarmoWAM reaches 95\% and 75\%, while the best baseline reaches 70\% and 60\%. In \textit{Put Flowers in Vase}, HarmoWAM achieves 95\%, 85\%, and 60\%, whereas the best baseline only reaches 80\%, 65\%, and 35\%. 
This further demonstrates that our approach can effectively harness the world model’s learned semantic priors, enabling the two action experts to be coordinated through explicit and implicit conditioning for reliable object approach and interaction with unseen objects.
This demonstrates that HarmoWAM can maintain generalizable transit execution under visual and spatial shifts while ensuring precise interaction during contact-intensive manipulation.

\input{tables/ablation_noise_step}
\subsection{Additional Ablation}
\label{app:ablation}
\begin{figure}[htbp] 
    \centering
    \includegraphics[width=\textwidth]{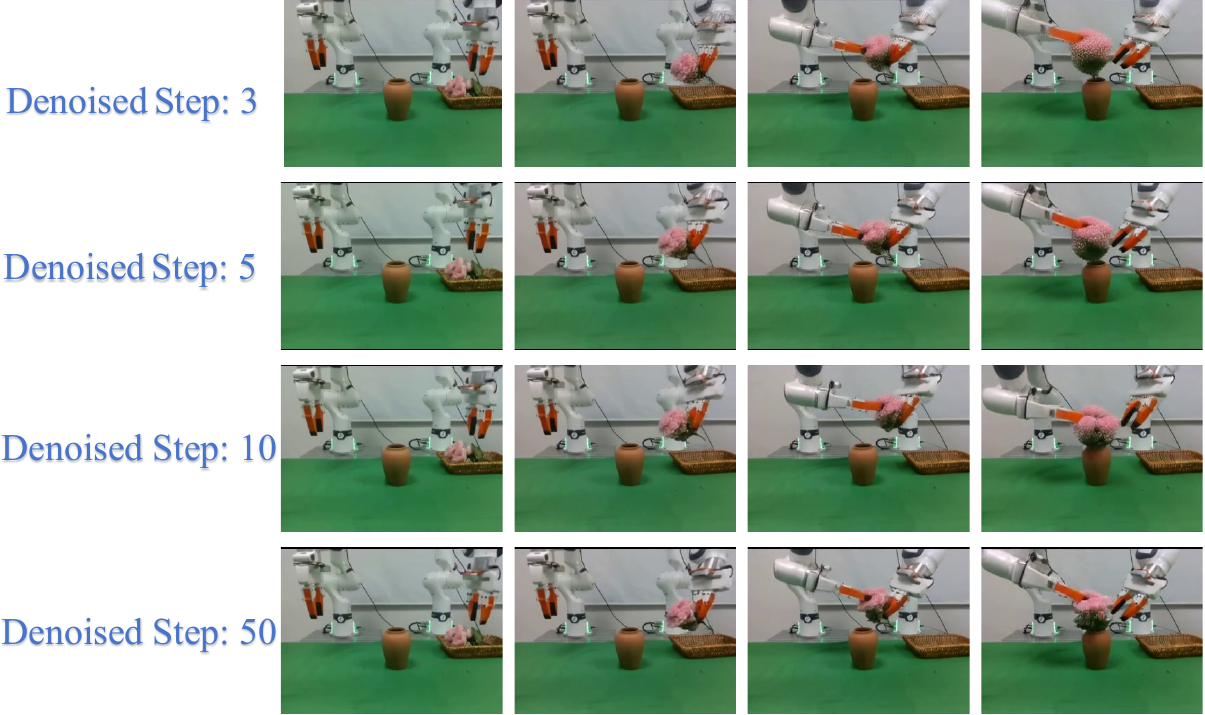} 
    \caption{Visual comparison of generated videos under different denoising steps.}
    \label{fig:denoising_steps_visualization}
    \vspace{-0.1cm}
\end{figure}

We ablate the number of denoising steps in the world model on the \textit{Put Flowers in Vase} task, using 3, 5, 10, and 50 steps, and compare their task success rates and inference frequencies. This study aims to identify the best trade-off between video prediction quality and generation efficiency. As shown in Table~\ref{tab:denoising_steps_ablation}, using 3 steps results in a lower success rate (80\%) due to coarse and blurry video predictions, while 5 steps lead to an improved success rate of 85\% without reducing inference speed. Increasing to 10 or 50 steps does not significantly improve success rates (85\% and 87\%, respectively) but decreases inference frequency, indicating diminishing returns for higher denoising steps. We further visualize the generated videos in Figure~\ref{fig:denoising_steps_visualization}: videos with 3 steps are noticeably blurred and lack key details, while 5-step videos preserve clarity of crucial action stages and object locations, ensuring reliable observation and task execution.

\section{Failure Case Analysis}
\label{app:failure}

\begin{figure}[htbp]
    \centering
    \includegraphics[width=\columnwidth]{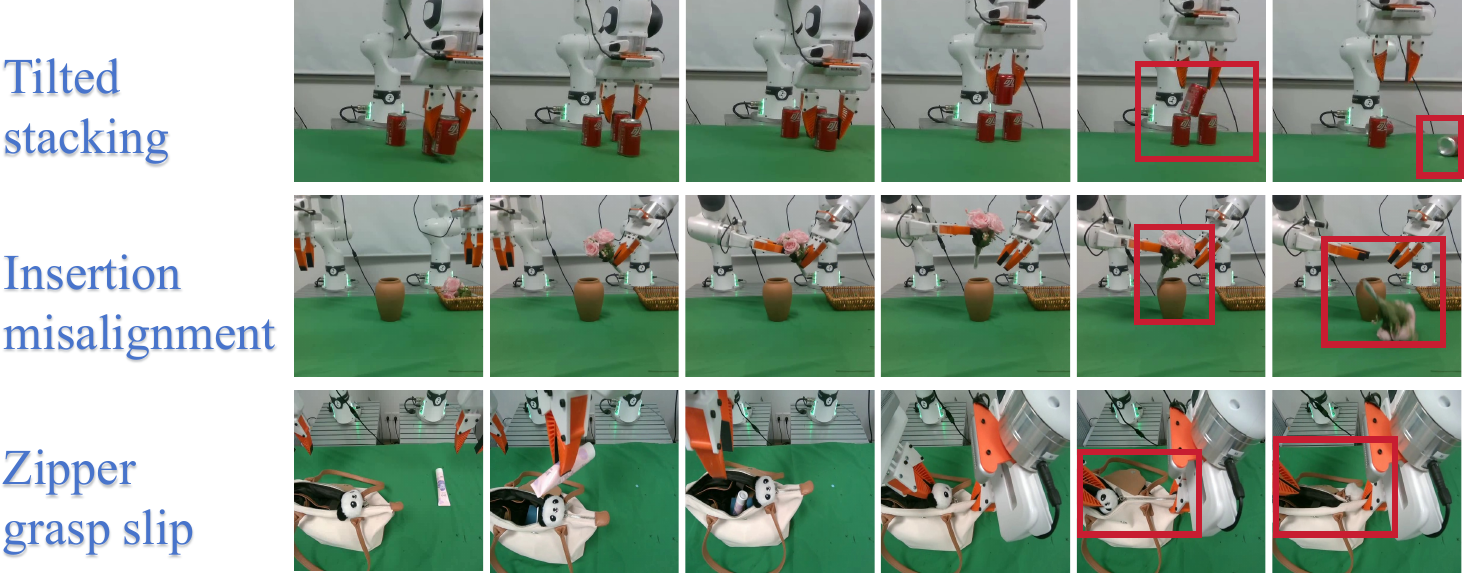}
    \vspace{-0.5cm} 
    \caption{\textbf{Failure case visualization of HarmoWAM.} We visualize representative failures in real-world Franka experiments, with red boxes highlighting the key error regions.}
    \label{fig:failure_cases}
    \vspace{-0.3cm} 
\end{figure}

Through real-world experiments on the Franka platform, we observed three specific failure cases encountered when HarmoWAM controlled the robot, as shown in Figure~\ref{fig:failure_cases}. Red boxes highlight the key errors.

\textbf{Tilted Stacking.}
In stacking tasks, HarmoWAM can usually complete the grasping and lifting motions, but the final placement may become unstable. The upper object may be placed with a slight front-back offset relative to the supporting object, causing it to tilt or slip after release. This failure mainly results from inaccurate front-back displacement estimation from the third-view camera, where small offsets along this direction are difficult to capture reliably. Although HarmoWAM can preserve the overall stacking intention and approach trajectory, stable stacking still requires highly accurate relative placement, where minor visual localization errors can lead to post-release instability.

\textbf{Insertion Misalignment.} 
In the flower insertion task, the handover and transport stages can succeed, but the final insertion stage may still fail as the stem brushes against the vase rim. This failure is mainly caused by slight variations in the grasping position, which change the effective exposed length of the stem and make the insertion pose harder to align. Since this task requires bimanual handover, accurate relative pose recovery, and tight-tolerance insertion, it remains challenging for all evaluated models. Even when HarmoWAM maintains the correct task progress and moves the flower toward the vase, small grasp-induced pose deviations can prevent smooth insertion into the vase opening.

\textbf{Zipper Grasp Slip.} 
During the zipper-pulling process, HarmoWAM may fail when the zipper slips out of the gripper. This failure is mainly caused by the material and mechanical properties of the gripper, which may not provide sufficient friction or stable contact to hold the small zipper pull during continuous pulling. Since zipper manipulation requires sustained force along a constrained direction, even a slight loss of contact can break the pulling process and prevent successful closure.

\section{Broader Impact}
\label{app:Broader Impact}
HarmoWAM aims to improve robotic manipulation by leveraging world-model-driven physical reasoning to coordinate predictive and reactive control. By enabling robust transit and precise interaction in unseen environments, our method reduces reliance on task-specific data collection and environment-specific retraining, supporting robot deployment beyond controlled laboratories.
This capability can benefit applications such as smart manufacturing, logistics automation, assistive robotics, and domestic service, where robots must handle diverse backgrounds, changing object positions, and novel object semantics. However, deploying robotic policies in physical environments may introduce risks, including unexpected behaviors caused by inaccurate predictions, out-of-distribution observations, or ambiguous instructions.
Therefore, real-world deployment should incorporate safety guardrails, uncertainty-aware control, human supervision, and hardware-level constraints. Overall, HarmoWAM contributes to more reliable robotic manipulation, helping robots move from laboratory demonstrations to complex, unstructured real-world environments.

\section{Limitations}
\label{app:limitation}
Since the pretrained world model in HarmoWAM operates with a fixed generation horizon, downstream tasks must preserve the same future video horizon to maintain alignment with its learned spatio-temporal dynamics.
While this is a common constraint in pretrained generative world models~\cite{wan2025wan, kim2026cosmos}, it limits the flexibility of adapting the prediction horizon to tasks with different temporal requirements. Future work will explore adaptive future-frame generation, where the prediction window can be dynamically adjusted based on task context and execution progress. Moreover, while pixel-level future generation provides intuitive visual guidance, it may introduce unnecessary generation overhead. Future work will further investigate latent-level predictive representations, enabling more efficient downstream action generation.

%% file: tables/pretrain.tex
\begin{table}[t]
\centering
\caption{
Statistics of the public robot datasets used for Wan2.2-TI2V-5B pretraining.
}
\label{tab:dataset_statistics}
\resizebox{0.95\linewidth}{!}{
\begin{tabular}{c|c|c}
\hline
\textbf{Dataset} & \textbf{Robot Arm / Platform} & \textbf{Number of Trajectories} \\
\hline
DROID & Franka Panda & 201,119  \\
AgiBot & AgiBot G1 & 3,017  \\
RoboMIND & Franka / UR / Ark / Agilex / TienKung & 1,721,985  \\
\hline
\end{tabular}
}
\end{table}

%% file: tables/experiments_in.tex
\begin{table}[t]
  \centering
  \caption{Stage-wise success rates for ID single-arm manipulation tasks.}
  \label{tab:in_domain}

  \footnotesize 
  \setlength{\tabcolsep}{0pt} 
  \renewcommand{\arraystretch}{1.4} 

  \begin{tabular*}{\textwidth}{@{\extracolsep{\fill}} l c c c c c}
    \toprule
    \multirow{2}{*}{\textbf{Method}} 
    & \textbf{Pick Fruit to Plate} 
    & \textbf{Stack Coke Cans} 
    & \textbf{Pour Coke into Beaker} 
    & \textbf{Write "Yes"} 
    & \multirow{2}{*}{\textbf{Avg}} \\
    \cmidrule{2-2} \cmidrule{3-3} \cmidrule{4-4} \cmidrule{5-5}
  
    & S1 $\rightarrow$ S2 
    & S1 $\rightarrow$ S2 
    & S1 $\rightarrow$ S2 
    & S1 $\rightarrow$ S2 $\rightarrow$ S3 
    & \\
    \midrule

    $\pi_{0.5}$ 
    & 0.85 \phantom{$\rightarrow$} 0.75 
    & 0.75 \phantom{$\rightarrow$} 0.60 
    & 0.80 \phantom{$\rightarrow$} 0.70
    & 0.90 \phantom{$\rightarrow$} 0.80 \phantom{$\rightarrow$} 0.80 
    & 0.77 \\

    VPP 
    & 0.85 \phantom{$\rightarrow$} 0.75
    & 0.65 \phantom{$\rightarrow$} 0.55 
    & 0.80 \phantom{$\rightarrow$} 0.75
    & 0.85 \phantom{$\rightarrow$} 0.75 \phantom{$\rightarrow$} 0.60
    & 0.73 \\

    Wan+Anypos 
    & 0.90 \phantom{$\rightarrow$} 0.85 
    & 0.85 \phantom{$\rightarrow$} 0.35
    & \textbf{0.90} \phantom{$\rightarrow$} 0.65 
    & 0.90 \phantom{$\rightarrow$} 0.70 \phantom{$\rightarrow$} 0.55
    & 0.74 \\

    QwenVLA-OFT 
    & 0.80 \phantom{$\rightarrow$} 0.75 
    & 0.35 \phantom{$\rightarrow$} 0.25 
    & 0.75 \phantom{$\rightarrow$} 0.70 
    & 0.80 \phantom{$\rightarrow$} 0.75 \phantom{$\rightarrow$} 0.60 
    & 0.64 \\

    Cosmos-Policy 
    & \textbf{0.95} \phantom{$\rightarrow$} 0.90
    & 0.75 \phantom{$\rightarrow$} 0.55
    & 0.85 \phantom{$\rightarrow$} 0.75
    & \textbf{0.95} \phantom{$\rightarrow$} 0.85 \phantom{$\rightarrow$} 0.70 
    & 0.81 \\
  
    \midrule
    \textbf{Ours} 
    & \textbf{0.95} \phantom{$\rightarrow$} \textbf{0.95} 
    & \textbf{0.95} \phantom{$\rightarrow$} \textbf{0.85} 
    & \textbf{0.90} \phantom{$\rightarrow$} \textbf{0.85} 
    & \textbf{0.95} \phantom{$\rightarrow$} \textbf{0.90} \phantom{$\rightarrow$} \textbf{0.90} 
    & \textbf{0.91} \\
    \bottomrule
  \end{tabular*}
\end{table}

%% file: tables/experiments_in_dual.tex
\begin{table*}[!t]
  \centering
  \caption{Stage-wise success rates for ID dual-arm manipulation tasks.}
  \label{tab:in_domain_dual}
  
  \footnotesize 
  \setlength{\tabcolsep}{0pt} 
  \renewcommand{\arraystretch}{1.4}

  \begin{tabular*}{\columnwidth}{@{\extracolsep{\fill}} l c c c}
    \toprule
    \multirow{2}{*}{\textbf{Method}} 
    & \textbf{Put Flowers to Vase} 
    & \textbf{Put Items to Bag and Zip} 
    & \multirow{2}{*}{\textbf{Avg}} \\
    \cmidrule{2-2} \cmidrule{3-3}
    
    & S1 $\to$ S2 $\to$ S3
    & S1 $\to$ S2 $\to$ S3 $\to$ S4 $\to$ S5
    & \\
    \midrule

    $\pi_{0.5}$ 
    & \textbf{0.95} \phantom{$\to$} 0.80 \phantom{$\to$} 0.40
    & 0.85 \phantom{$\to$} 0.75 \phantom{$\to$} 0.70 \phantom{$\to$} 0.55 \phantom{$\to$} 0.50
    & 0.69 \\

    Wan+Anypos 
    & 0.85 \phantom{$\to$} 0.60 \phantom{$\to$} 0.15
    & 0.75 \phantom{$\to$} 0.60 \phantom{$\to$} 0.45 \phantom{$\to$} 0.40 \phantom{$\to$} 0.40
    & 0.53 \\

    Cosmos-Policy 
    & \textbf{0.95} \phantom{$\to$} \textbf{0.95} \phantom{$\to$} 0.35
    & \textbf{0.90} \phantom{$\to$} 0.80 \phantom{$\to$} 0.70 \phantom{$\to$} 0.65 \phantom{$\to$} 0.55
    & 0.73 \\

    \midrule
    \textbf{Ours} 
    & \textbf{0.95} \phantom{$\to$} 0.90 \phantom{$\to$} \textbf{0.70}
    & \textbf{0.90} \phantom{$\to$} \textbf{0.90} \phantom{$\to$} \textbf{0.85} \phantom{$\to$} \textbf{0.80} \phantom{$\to$} \textbf{0.80}
    & \textbf{0.85} \\
    \bottomrule
  \end{tabular*}
\end{table*}

%% file: tables/experiments_out.tex
\begin{table}[t]
  \centering
  \caption{Stage-wise success rates for OOD single-arm manipulation tasks.}
  \label{tab:ood}
  
  \footnotesize 
  \setlength{\tabcolsep}{0pt} 
  \renewcommand{\arraystretch}{1.3} 

  \begin{tabular*}{\textwidth}{@{\extracolsep{\fill}} l l c c c c c}
    \toprule
    \multirow{2}{*}{\textbf{Method}} & \multirow{2}{*}{\textbf{Gen Scenario}}
    & \textbf{Pick Fruit} 
    & \textbf{Stack Cans} 
    & \textbf{Pour Coke} 
    & \textbf{Write "Yes"} 
    & \multirow{2}{*}{\textbf{Avg}} \\
    \cmidrule{3-3} \cmidrule{4-4} \cmidrule{5-5} \cmidrule{6-6}
    & 
    & S1 $\to$ S2 
    & S1 $\to$ S2 
    & S1 $\to$ S2 
    & S1 $\to$ S2 $\to$ S3 
    & \\
    \midrule

    \multirow{3}{*}{$\pi_{0.5}$}
    & Background & 0.75 \phantom{$\to$} 0.55 & 0.45 \phantom{$\to$} 0.35 & 0.70 \phantom{$\to$} 0.55 & 0.80 \phantom{$\to$} 0.65 \phantom{$\to$} 0.55 & 0.59 \\
    & Position   & 0.25 \phantom{$\to$} 0.15 & 0.10 \phantom{$\to$} 0.05 & 0.45 \phantom{$\to$} 0.40 & 0.60 \phantom{$\to$} 0.55 \phantom{$\to$} 0.50 & 0.34 \\
    & Objects   & 0.65 \phantom{$\to$} 0.45 & 0.50 \phantom{$\to$} 0.35 & 0.55 \phantom{$\to$} 0.50 & 0.85 \phantom{$\to$} 0.60 \phantom{$\to$} 0.45 & 0.54 \\
    \midrule

    \multirow{3}{*}{VPP}
    & Background & 0.50 \phantom{$\to$} 0.25 & 0.35 \phantom{$\to$} 0.15 & 0.70 \phantom{$\to$} 0.65 & 0.55 \phantom{$\to$} 0.40 \phantom{$\to$} 0.25 & 0.42 \\
    & Position   & 0.30 \phantom{$\to$} 0.25 & 0.15 \phantom{$\to$} 0.05 & 0.25 \phantom{$\to$} 0.15 & 0.50 \phantom{$\to$} 0.35 \phantom{$\to$} 0.15 & 0.24 \\
    & Objects   & 0.80 \phantom{$\to$} 0.55 & 0.55 \phantom{$\to$} 0.25 & 0.70 \phantom{$\to$} 0.65 & 0.70 \phantom{$\to$} 0.45 \phantom{$\to$} 0.35 & 0.56 \\
    \midrule

    \multirow{3}{*}{Wan+Anypos}
    & Background & 0.60 \phantom{$\to$} 0.55 & 0.60 \phantom{$\to$} 0.25 & 0.80 \phantom{$\to$} 0.50 & 0.80 \phantom{$\to$} 0.60 \phantom{$\to$} 0.55 & 0.58 \\
    & Position   & 0.85 \phantom{$\to$} 0.80 & 0.55 \phantom{$\to$} 0.25 & 0.35 \phantom{$\to$} 0.30 & 0.75 \phantom{$\to$} 0.40 \phantom{$\to$} 0.35 & 0.51 \\
    & Objects   & 0.80 \phantom{$\to$} 0.45 & 0.70 \phantom{$\to$} 0.60 & 0.75 \phantom{$\to$} 0.60 & 0.80 \phantom{$\to$} 0.70 \phantom{$\to$} 0.55 & 0.66 \\
    \midrule

    \multirow{3}{*}{QwenVLA-OFT}
    & Background & 0.70 \phantom{$\to$} 0.25 & 0.15 \phantom{$\to$} 0.10 & 0.60 \phantom{$\to$} 0.55 & 0.70 \phantom{$\to$} 0.60 \phantom{$\to$} 0.60 & 0.47 \\
    & Position   & 0.45 \phantom{$\to$} 0.10 & 0.15 \phantom{$\to$} 0.05 & 0.40 \phantom{$\to$} 0.25 & 0.55 \phantom{$\to$} 0.40 \phantom{$\to$} 0.30 & 0.29 \\
    & Objects   & 0.65 \phantom{$\to$} 0.60 & 0.15 \phantom{$\to$} 0.05 & 0.75 \phantom{$\to$} 0.70 & 0.60 \phantom{$\to$} 0.50 \phantom{$\to$} 0.45 & 0.49 \\
    \midrule

    \multirow{3}{*}{Cosmos-Policy}
    & Background & \textbf{0.90} \phantom{$\to$} 0.80 & 0.60 \phantom{$\to$} 0.15 & 0.80 \phantom{$\to$} 0.65 & 0.85 \phantom{$\to$} 0.45 \phantom{$\to$} 0.40 & 0.62 \\
    & Position   & 0.25 \phantom{$\to$} 0.15 & 0.30 \phantom{$\to$} 0.10 & 0.15 \phantom{$\to$} 0.15 & 0.65 \phantom{$\to$} 0.50 \phantom{$\to$} 0.50 & 0.31 \\
    & Objects   & 0.85 \phantom{$\to$} 0.75 & 0.60 \phantom{$\to$} 0.15 & 0.70 \phantom{$\to$} 0.65 & 0.35 \phantom{$\to$} 0.15 \phantom{$\to$} 0.15 & 0.48 \\
    \midrule

    \multirow{3}{*}{\textbf{Ours}}
    & Background & \textbf{0.90} \phantom{$\to$} \textbf{0.90} 
        & \textbf{0.90} \phantom{$\to$} \textbf{0.65} & \textbf{0.85} \phantom{$\to$} \textbf{0.80} & \textbf{0.90} \phantom{$\to$} \textbf{0.85} \phantom{$\to$} \textbf{0.70} & \textbf{0.83} \\
    & Position   & \textbf{0.95} \phantom{$\to$} \textbf{0.90} 
        & \textbf{0.85} \phantom{$\to$} \textbf{0.70} & \textbf{0.85} \phantom{$\to$} \textbf{0.75} & \textbf{0.85} \phantom{$\to$} \textbf{0.85} \phantom{$\to$} \textbf{0.80} & \textbf{0.83} \\
    & Objects   & \textbf{0.95} \phantom{$\to$} \textbf{0.80} & \textbf{0.95} \phantom{$\to$} \textbf{0.75} & \textbf{0.90} \phantom{$\to$} \textbf{0.85} & \textbf{0.90} \phantom{$\to$} \textbf{0.85} \phantom{$\to$} \textbf{0.85} & \textbf{0.87} \\
    \bottomrule
  \end{tabular*}
\end{table}

%% file: tables/experiments_out_dual.tex
\begin{table}[htbp]
  \centering
  \caption{Stage-wise success rates for OOD dual-arm manipulation tasks.}
  \label{tab:ood_dual}
  
  \footnotesize 
  \setlength{\tabcolsep}{0pt} 
  \renewcommand{\arraystretch}{1.4} 

  \begin{tabular*}{\columnwidth}{@{\extracolsep{\fill}} l l c c c}
    \toprule
    \multirow{2}{*}{\textbf{Method}} & \multirow{2}{*}{\textbf{Gen Scenario}} & \textbf{Put Flowers to Vase} & \textbf{Put Items to Bag and Zip} & \multirow{2}{*}{\textbf{Avg}} \\
    \cmidrule{3-3} \cmidrule{4-4}
    & & S1 $\to$ S2 $\to$ S3 & S1 $\to$ S2 $\to$ S3 $\to$ S4 $\to$ S5 & \\
    \midrule

    \multirow{3}{*}{$\pi_{0.5}$} 
    & Background & \textbf{0.90}  \phantom{$\to$} 0.65 \phantom{$\to$} 0.30 & 0.70 \phantom{$\to$} 0.65 \phantom{$\to$} 0.65 \phantom{$\to$} 0.50 \phantom{$\to$} 0.50 & 0.61 \\
    & Position   & 0.50 \phantom{$\to$} 0.35 \phantom{$\to$} 0.25 & 0.35 \phantom{$\to$} 0.30 \phantom{$\to$} 0.25 \phantom{$\to$} 0.20 \phantom{$\to$} 0.20 & 0.30 \\
    & Objects   & 0.80 \phantom{$\to$} 0.65 \phantom{$\to$} 0.35 & 0.65 \phantom{$\to$} 0.55 \phantom{$\to$} 0.45 \phantom{$\to$} 0.40 \phantom{$\to$} 0.40 & 0.53 \\
    \midrule

    \multirow{3}{*}{Wan+Anypos} 
    & Background & 0.75 \phantom{$\to$} 0.55 \phantom{$\to$} 0.10 & 0.45 \phantom{$\to$} 0.40 \phantom{$\to$} 0.40 \phantom{$\to$} 0.35 \phantom{$\to$} 0.30 & 0.41 \\
    & Position   & 0.70 \phantom{$\to$} 0.50 \phantom{$\to$} 0.15 & 0.60 \phantom{$\to$} 0.55 \phantom{$\to$} 0.40 \phantom{$\to$} 0.35 \phantom{$\to$} 0.35 & 0.45 \\
    & Objects   & 0.65 \phantom{$\to$} 0.45 \phantom{$\to$} 0.10 & 0.50 \phantom{$\to$} 0.45 \phantom{$\to$} 0.45 \phantom{$\to$} 0.40 \phantom{$\to$} 0.35 & 0.42 \\
    \midrule

    \multirow{3}{*}{Cosmos-Policy} 
    & Background & 0.65 \phantom{$\to$} 0.45 \phantom{$\to$} 0.15 & 0.65 \phantom{$\to$} 0.50 \phantom{$\to$} 0.50 \phantom{$\to$} 0.40 \phantom{$\to$} 0.35 & 0.46 \\
    & Position   & 0.35 \phantom{$\to$} 0.30 \phantom{$\to$} 0.10 & 0.35 \phantom{$\to$} 0.25 \phantom{$\to$} 0.25 \phantom{$\to$} 0.10 \phantom{$\to$} 0.10 & 0.23 \\
    & Objects   & 0.80 \phantom{$\to$} 0.65 \phantom{$\to$} 0.05 & 0.70 \phantom{$\to$} 0.55 \phantom{$\to$} 0.40 \phantom{$\to$} 0.35 \phantom{$\to$} 0.20 & 0.46 \\
    \midrule

    \multirow{3}{*}{\textbf{Ours}} 
    & Background & 0.85 \phantom{$\to$} \textbf{0.75} \phantom{$\to$} \textbf{0.60} & \textbf{0.85} \phantom{$\to$} \textbf{0.85} \phantom{$\to$} \textbf{0.80} \phantom{$\to$} \textbf{0.70} \phantom{$\to$} \textbf{0.70} & \textbf{0.76} \\
    & Position   & \textbf{0.90} \phantom{$\to$} \textbf{0.85} \phantom{$\to$} \textbf{0.65} & \textbf{0.70} \phantom{$\to$} \textbf{0.65} \phantom{$\to$} \textbf{0.65} \phantom{$\to$} \textbf{0.60} \phantom{$\to$} \textbf{0.60} & \textbf{0.70} \\
    & Objects   & \textbf{0.95} \phantom{$\to$} \textbf{0.85} \phantom{$\to$} \textbf{0.60} & \textbf{0.85} \phantom{$\to$} \textbf{0.85} \phantom{$\to$} \textbf{0.80} \phantom{$\to$} \textbf{0.80} \phantom{$\to$} \textbf{0.75} & \textbf{0.81} \\
    \bottomrule
  \end{tabular*}
\end{table}

%% file: tables/ablation_noise_step.tex
\begin{table}[t]
\centering
\caption{
Ablation study on the number of denoising steps in System 2. 
}
\label{tab:denoising_steps_ablation}
\resizebox{0.65\linewidth}{!}{
\begin{tabular}{c|c|c}
\hline
\textbf{Denoising Steps} & \textbf{Success Rate (\%)} & \textbf{Inference Frequency (Hz)} \\
\hline
3  & 80 & 4 \\
\textbf{5}  & \textbf{85} & \textbf{4} \\
10 & 85 & 3.6 \\
50 & 87 & 3 \\
\hline
\end{tabular}
}
\end{table}